\documentclass{article}

\usepackage{PRIMEarxiv}

\usepackage[utf8]{inputenc} 
\usepackage[T1]{fontenc}    
\usepackage{hyperref}       
\usepackage{url}            
\usepackage{booktabs}       
\usepackage{amsfonts}       
\usepackage{nicefrac}       
\usepackage{microtype}      
\usepackage{lipsum}
\usepackage{fancyhdr}       
\usepackage{graphicx}       
\usepackage{comment}
\usepackage{natbib}
\graphicspath{{media/}}     

\usepackage[utf8]{inputenc} 
\usepackage[T1]{fontenc}    
\usepackage{hyperref}       
\usepackage{url}            
\usepackage{booktabs}       
\usepackage{amsfonts}       
\usepackage{nicefrac}       
\usepackage{microtype}      
\usepackage{xcolor}         

\usepackage{amsmath}
\usepackage{algorithm}
\usepackage{mathtools}
\usepackage{mathrsfs}
\usepackage{multirow}
\usepackage{amsfonts}
\usepackage{amsmath}
\usepackage{amsthm}
\usepackage{color}
\usepackage{mathtools}
\usepackage{algorithm}
\usepackage{algpseudocode}
\usepackage{amssymb}
\usepackage[mathscr]{euscript}
\usepackage[toc]{appendix}
\usepackage{comment}

\newcommand{\indep}{\perp \!\!\! \perp}

\usepackage{graphicx}
\usepackage{float}
\usepackage{subfig}
\usepackage{enumitem}
\newtheorem{assumption}{Assumption}

\newtheorem{definition}{Definition}

\newtheorem{lemma}{Lemma}
\newtheorem{proposition}{Proposition}
\newtheorem{theorem}{Theorem}

\title{Conformal Off-Policy Prediction
}

\author{
  Yingying Zhang \\
  Academy of Statistics and Interdisciplinary Sciences, KLATASDS-MOE\\
  East China Normal University\\
  Shanghai, China\\
   \And
  Chengchun Shi \\
  Department of Statistics \\
  London School of Economics and Political Science \\
  London, United Kingdom\\
    \AND
  Shikai Luo \\
  Bytedance \\
  Beijing, China
}

\begin{document}

	\maketitle

\begin{abstract}
		Off-policy evaluation is critical in a number of applications where new policies need to be evaluated offline before online deployment. Most existing methods focus on the expected return, define the target parameter through averaging and provide a point estimator only. In this paper, we develop a novel procedure to produce reliable interval estimators for a target policy's return starting from any initial state. Our proposal accounts for the variability of the return around its expectation, focuses on the individual effect and offers valid uncertainty quantification. Our main idea lies in designing a pseudo policy that generates subsamples as if they were sampled from the target policy so that existing conformal prediction algorithms are applicable to prediction interval construction. Our methods are justified by theories, synthetic data and real data from short-video platforms. 
	\end{abstract}
	
	\section{Introduction}
	
	Policy evaluation plays a crucial role in many real-world applications including healthcare,  marketing, social sciences, among many others. Before deploying any new policy, it is crucial to know the impact of this policy. However, in the aforementioned applications, it is often impractical 
	to evaluate a new policy by directly running this policy. As a result, the new policy needs to be evaluated offline based on an observational dataset generated by a possibly different behavior policy. This formulates the off-policy evaluation (OPE) problem. 
	
	Most works in the literature focus on evaluating the \textit{average} value of a target policy aggregated over different initial states. In many applications such as healthcare and technology industries, in addition to the average effect, it is crucial to learn the value under a given initial condition (e.g., the individual effect) as well. For instance, in precision medicine, it allows us to estimate the outcome of each individual patient following a given treatment regime. In online recommendation, it allows us to evaluate the effect of a new strategy for each individual visitor. Moreover, real world datasets often follow asymmetric or heavy-tailed distributions. An example is given in our real dataset collected from a world-leading short-video platform where the outcome distribution is highly heavy-tailed (see Figure \ref{LLP}). 
	In these applications, in additional to a target policy's \textit{mean} return, it is equally important to infer its outcome distribution. This motivates us to construct a prediction interval for the target policy's outcome.  

	\subsection{Related Work}	
	\textbf{Off-policy evaluation}. There is a huge literature on OPE. Existing methods can be divided into three categories, corresponding to the value-based method \citep[see e.g.,][]{le2019batch,luckett2020estimating,liao2021off,chen2022well,shi2020statistical}, importance sampling (IS) or resampling-based method \citep[see e.g.,][]{precup2000eligibility,li2011unbiased,liu2018breaking,nachum2019dualdice,schlegel2019importance} and doubly robust method \citep[see e.g.,][]{farajtabar2018more,kallus2018policy,tang2019doubly,uehara2020minimax,kallus2020double,liao2022batch}. 
	
	In addition, several papers have studied interval estimation of the policy's value for uncertainty quantification  \citep{thomas2015high,jiang2016doubly,hanna2017bootstrapping,dai2020coindice,feng2020accountable,jiang2020minimax,chandak2021universal,hao2021bootstrapping,shi2021deeply,wang2021projected}. These confidence intervals are typically derived based on concentration inequalities, normal approximations, bootstrap \citep{efron1994introduction} or the empirical likelihood method \citep{owen2001empirical}. However, all the aforementioned methods focused on the average effect of the target policy. To our knowledge, interval estimation of the individual effect has been less explored in the literature. 
	
	\textbf{Conformal prediction}. 
	Our proposal is closely related to a line of research on conformal prediction (CP), which was originally introduced by \cite{vovk2005conformal} to construct valid model-free \textit{prediction} intervals (PIs) for the response; see also, some follow-up works by \citet{vovk2009line,vovk2012conditional,lei2014distribution,lei2015conformal,lei2018distribution,sesia2020comparison,cauchois2021knowing}. Both PI and confidence interval (CI) express uncertainty in statistical estimates. Nonetheless, a CI gives a range for the conditional mean function of the response whereas a PI aims to cover the response itself. 
	A key strength of CP lies in its generality and finite-sample guarantees. Specifically, it can accommodate \textit{any} prediction model under minimal assumptions on the data and achieve nominal coverage even for small samples.
	
	Recently, \cite{tibshirani2019conformal} developed a weighted CP method to handle settings under covariate shift. The weighted CP method was further extended and applied to a number of applications, including individual treatment effects estimation \citep{kivaranovic2020conformal,jin2021sensitivity,lei2021conformal,yin2022conformal}, survival analysis \cite{candes2021conformalized}, classification under covariate shift  \citep{podkopaev2021distribution}, to name a few. These methods considered either a standard supervised learning setting, or a contextual bandit setting with ``state-agnostic" target policies. These settings differ from ours that involve sequential decision making and a general target policy. 
	
	Finally, we notice that there is a closely related concurrent work by \citet{taufiq2022conformal} that studied conformal off-policy prediction in contextual bandits. However, they did not consider sequential decision making, which is more challenging. In addition, even when specialized to contextual bandits, the proposed methodology differs largely from theirs. See Section \ref{sec:challenge} for more details. 

\textbf{Distributional reinforcement learning}. Recently, there is an emerging line of research on distributional reinforcement learning that estimates the entire distribution of the return under the optimal policy \citep[see e.g.,][]{bellemare2017distributional,dabney2018distributional,mavrin2019distributional,zhou2020non}. Our proposal shares similar spirits with these works in that it not only considers the expected return, but takes the variability of the return around its expectation into account as well.

\subsection{Contribution}
Methodologically, we develop a novel procedure to construct off-policy PIs for a target policy's return starting from any initial state in sequential decision making. 
It is ultimately different from many existing OPE methods that consider the average effect aggregated over different initial states, construct CIs for the expected return and ignore the variance of the return around its expectation. A key ingredient of our proposal lies in constructing a pseudo variable whose distribution depends on both the target and behavior policy. We next sample a subset of observations based on this pseudo variable and apply weighted conformal prediction method on the selected subsamples. Finally, we develop an importance-sampling-based method and a multi-sampling-based method to further improve efficiency. 

Theoretically, we prove that the proposed PI achieves valid coverage asymptotically. In addition, when the behavior policy is known to us (e.g., as in randomized studies), it achieves exact coverage in finite samples. Such a property is particularly appealing as the sample size is usually limited in offline domains. Finally, our PI is asymptotically efficient when the regression estimator is consistent.  

\section{Preliminaries: Conformal Prediction}\label{sec:2}
We begin with a brief overview for the CP algorithm in 
supervised learning. Given i.i.d. predictor-response pairs $\{Z_i=(X_i, Y_i)\}_{i=1}^n$, it is concerned with producing a prediction band $\widehat{C}(\bullet)$ (as a function of the predictor $X_i$) such that
for an identically distributed test data pair $Z_{n+1}=(X_{n+1},Y_{n+1})$, 
\begin{eqnarray}\label{eqn:coverage}
	\mathbb{P}(Y_{n+1}\in \widehat{C}(X_{n+1}))\ge 1-\alpha,
\end{eqnarray}
for a given desired coverage rate $1-\alpha\in (0,1)$. One example of $\widehat{C}(x)$ is given by $[q_{\alpha_{L}}(x), q_{\alpha_{U}}(x)]$ where $q_{\alpha_{L}}(x)$ and $q_{\alpha_{U}}(x)$ correspond to the $\alpha_L$th (lower) and $\alpha_U$th (upper) conditional quantiles of $Y$ given $X=x$  such that $\alpha_U-\alpha_L=1-\alpha$. Given the observed data, we can employ state-of-the-art machine learning algorithms to learn these conditional quantiles. This yields the following $\widehat{C}(x)$, 
\begin{eqnarray}\label{eqn:PI}
	\widehat{C}(x)=[\widehat{q}_{\alpha_{L}}(x), \widehat{q}_{\alpha_{U}}(x)].	
\end{eqnarray}
However, one disadvantage of the aforementioned PI is that the inequality \eqref{eqn:coverage} is not guaranteed to hold in finite samples, due to the estimation errors of the the conditional quantiles. 

The CP algorithm is developed to address this challenge. 
At a high-level, CP allows us to calibrate PIs (such as \eqref{eqn:PI}) computed by general black box machine learning algorithms with finite-sample guarantees such that \eqref{eqn:coverage} holds for any $n$. 
Specifically, CP first splits the data into training and calibration data-subsets. On the training dataset, it learns the conditional mean or quantile of $Y$ given $X$ using any black box machine learning algorithm.  For instance, let $\widehat{\mu}$ denote the estimated conditional mean function. On the calibration dataset $\{(X_i^{cal}, Y_i^{cal})\}_{i}$, 
it calculates a nonconformity score (e.g., $|Y_i^{cal}-\widehat{\mu}(X_i^{cal})|$) 
that measures how each observation ``conforms" to the training dataset. The resulting PI is constructed based on the empirical quantiles of these nonconformity scores and attains valid coverage as long as the data observations are exchangeable. There are many choices of the score function available and we refer 
readers to  \cite{gupta2021nested} for details. Another widely-used score function 
is given by 
\begin{eqnarray}\label{eqn:score}
	\max\{\widehat{q}_{\alpha_{L}}(X_i^{cal})-Y_i^{cal}, Y_i^{cal}-\widehat{q}_{\alpha_{U}}(X_i^{cal})\}
\end{eqnarray} 
where $\widehat{q}_{\alpha_L}$ and $\widehat{q}_{\alpha_U}$ are the conditional quantile estimators given in \eqref{eqn:PI}. 
The resulting algorithm is referred to as the conformal quantile regression \citep{romano2019conformalized}.


We next review the weighted CP algorithm developed by \cite{tibshirani2019conformal}. As commented earlier, the aforementioned CP algorithm relies on exchangeability --- a key assumption that requires 
the joint distribution of calibration and testing samples to be invariant to the order of samples. This assumption 
is clearly violated under distributional shift where the calibration and testing samples follow different distribution functions. To address this concern, \cite{tibshirani2019conformal} introduced the so-called 
``weighted exchangeability'' that relaxes the classical i.i.d. assumption and is automatically satisfied for independent samples. 

\begin{definition}(Weighted Exchangeability)  $Z_1,\ldots,Z_n$ and the testing sample $Z_{n+1}$ are said to be weighted exchangeable,
	if the density $f$ of their joint distribution can be factorized as 
	\begin{equation*}
		f(z_1,\ldots,z_{n+1})=\prod_{i=1}^{n+1}w_i(z_i)g(z_1,\ldots,z_{n+1}),
	\end{equation*}
	for certain weight functions $\{w_i\}_{i=1}^{n+1}$, 
	and a permutation-invariance function $g$ such that $g(z_{\sigma(1)},\ldots,z_{\sigma(n+1)})=g(z_1,\ldots,z_{n+1})$ for any permutation $\sigma$ of $\{1,\ldots,n+1\}$.
\end{definition}
According to the definition, independent data are always ``weighted exchangeable" with weight function corresponding to the likelihood ratios.
\begin{lemma}\label{lemma1}
	Let $Z_i\sim P_i$, $i=1,\ldots,n+1$ be independent draws, where each $P_{i}$ is absolutely continuous with respect to $P_1$ for $i\ge 2$. Then $Z_1,\ldots,Z_{n+1}$ are weighted exchangeable with weight functions $w_1=1$ and $w_i=dP_i/dP_1$, $i\ge 2$.
\end{lemma}
Let $S_i=\mathcal{S}(Z_i^{cal},\mathcal{Z}^{tr})$ denote the nonconformity score for the $i$th observation in the calibration data based on certain machine learning algorithm trained on the training dataset $\mathcal{Z}^{tr}$, and $S_{(x,y)}=\mathcal{S}((x,y),\mathcal{Z}^{tr})$ denote the one for an arbitrary predictor-response pair $(x,y)$. Instead of relying on the empirical quantiles of these nonconformity scores, the weighted CP algorithm considers a weighted version and constructs the PI for $Y$ given $X=x$ 
as
\begin{eqnarray*}
	\{y: S_{(x,y)}\le (1-\alpha)\mbox{th~quantile~of~the~mixture~distribution}\\
	\sum_{i=1}^{n}p_{i}^{w}\delta_{S_{i}}+p_{n+1}^{w}\delta_{\infty} \}
\end{eqnarray*}
where $1-\alpha$ denotes the desired coverage rate, $\delta_{a}$ denotes the Dirac delta distribution that places all mass at the value $a$, and the mixing probabilities $\{p_{i}^{w}\}_{i=1}^{n+1}$ are functions of weights $\{w_{i}\}_{i=1}^{n+1}$ whose explicit expressions are given in
\citet{tibshirani2019conformal}.

Finally, we remark that the (weighted) CP method possesses several appealing statistical properties. First, it does not depend on any specific model assumption in the conditional distribution of the outcome given the covariates; as such, it is applicable to complex nonlinear and high-dimensional settings. Second, it achieves exact coverage in the sense that $\mathbb{P}\{Y \in \widehat{C}_n(X)\}\ge 1-\alpha$ for any $n$. To the contrary, most interval estimation procedures are only \textit{asymptotically} valid. Nonetheless, it is not straightforward to extend these methods to the OPE problem. See Section \ref{sec:challenge} for details. 

\section{Conformal Off-Policy Prediction in Contextual Bandits}
\label{sec:3}

\subsection{Problem Formulation}
To better illustrate the idea, in this section, we focus on a contextual bandit setting (i.e., single stage decision making) where the observed data consist of $n$ i.i.d. samples $\{(X_i,T_i,Y_i)\}_{i=1}^{n}$ where $X_i$ collects the contextual information of the $i$th instance, $T_i\in \{0,1,\cdots,m-1\}$ denotes the treatment (e.g., action) that the $i$th instance receives where $m$ denotes the number of treatment options, and $Y_i$ is the corresponding response (e.g., reward). We adopt a counterfactual/potential outcome framework \citep{rubin2005causal} to formulate the OPE problem. Specifically, for any $0\le t\le m-1$, let $Y_i^t$ denote the reward that the $i$th instance would have been observed were they to receive action $t$. 

A policy $\pi$ is a (stochastic) decision rule that maps the contextual space to a distribution function over the action space. We use $\pi(t|x)$ to denote the probability that the agent selects treatment $t$ given $X=x$. 
For a given target/evaluation policy $\pi_e$, we are interested in inferring the conditional distribution of the potential outcome $Y_{n+1}^{\pi_e}$ that would be observed were the instance to follow $\pi_{e}$ given $X_{n+1}$. Specifically, given $X_{n+1}$, we aim to produce a PI for $Y_{n+1}^{\pi_e}$ with valid coverage guarantees. Notice that our objective differs from the standard OPE problem in which one aims to derive a CI for the expected value $\mathbb{E} Y^{\pi_e}_{n+1}$. 

Finally, we impose standard assumptions in the causal inference literature \citep[see e.g.,][]{zhang2012robust,zhu2017greedy,chen2022policy}, including (1) $Y_i^{T_i}=Y_i$ almost surely for any $i$ (i.e., consistency); (2) $(Y_i^0, \cdots, Y_i^{m-1}) \indep T_i|X_i$ for any $i$ (i.e., no unmeasured confounders); (3) The behavior policy $\pi_b(t|x)=\mathbb{P}(T_i=t|X_i=x)$ is uniformly bounded away from zero for any $t,x$ (i.e., positivity).


\subsection{Conformal Prediction for Off-Policy Evaluation}\label{sec:challenge}
To motivate our proposed approach, we first outline two potential extensions of CP to the OPE problem in this section, corresponding to the direct method and the subsampling-based method, and discuss their limitations. We next illustrate the main idea of our proposal.

{\textbf{Direct method}}. OPE is essentially a policy evaluation problem under distribution shift where the target policy $\pi_e$  differs from the behavior policy $\pi_b$ that generates the offline data. By Lemma \ref{lemma1}, the calibration dataset $\{ (X_i^{cal},Y_i^{cal})\}_i$ and the predictor-potential outcome pair $(X_{n+1},Y_{n+1}^{\pi_e})$ in the target population satisfy weighed exchangeability with weights $1$ for samples in the calibration dataset and $w_{n+1}(x,y)$ (given below) for the testing data
\begin{eqnarray}\label{eqn:weight}
w_{n+1}(x,y)=\frac{dP_{Y^{\pi_e}|X}(y|x)}{dP_{Y|X}(y|x)},
\end{eqnarray}
where $P_{Y|X}$ and $P_{Y^{\pi_e}|X}$ denote the conditional distributions of $Y$ and $Y^{\pi_e}$ given $X$, respectively. 
As a result, a direct application of the weighted CP method is valid for OPE given the weights $\{w_i\}_{i}$. We refer to the resulting algorithm as the direct method and notice that the concurrent work by \citet{taufiq2022conformal} adopted a similar idea. 
One key step in their proposal is to use the estimated conditional density function and the Monte Carlo method to learn the weight function (see equation 7 in \citet{taufiq2022conformal}  for details). As such, their method can be sensitive to the specification of the conditional density model. On the contrary, our proposal below is robust to the model misspecification. We also conduct simulation studies in Section \ref{sec:5} to empirically verify the robustness property of our proposal.

To apply weighted CP, it remains to specify the weight $w_{n+1}$. Notice that both $Y$ and $Y^{\pi_e}$ correspond to a mixture of $\{Y^t: 1\le t\le m\}$ with different weight vectors. Estimating $w_{n+1}$ essentially requires to learn the conditional densities of $Y^t$ given $X$ --- an extremely challenging task in complicated high-dimensional nonlinear systems. As will show later, this approach would fail to cover $Y^{\pi_e}$ when the conditional density model is misspecified.

{\textbf{Subsampling-based method}}. 
Another approach to handle distributional shift is to take a data subset whose distribution is similar to the ``target distribution" and apply standard CP to this sub-dataset. In particular, for each observation $(X_i,T_i,Y_i)$ in the calibration dataset, we sample a pseudo action $E_i$ following the evaluation policy $\pi_e$, select subsamples whose pseudo action matches the observed action, and apply CP to these subsamples. We refer to the resulting algorithm as the subsampling-based method. 

However, this approach is not valid and is likely to produce PIs that undercover the target outcome $Y_{n+1}^{\pi_e}$ in general. This is because the distribution of the selected subsamples $\{(X_i,Y_i): T_i = E_i \}$ generally differs from that of $(X_{n+1}, Y_{n+1}^{\pi_e})$. The two distributions coincide only when $\pi_e$ is deterministic or $\pi_b$ is uniformly random, as shown below. 

\begin{proposition}\label{prop:1}
Let $E$ denote a pseudo action generated according to the target policy $\pi_e$. Then the conditional distribution of $Y$ given $E=T$ and $X$ follows a mixture distribution given as follows
\begin{equation*}
	P_{Y|E=T,X}=\sum_{t=0}^{m-1} \frac{\pi_e(t|X)\pi_b(t|X)}{\sum_{t'}\pi_e(t'|X)\pi_b(t'|X) }P_{Y^t|X}.
\end{equation*}
The above mixture distribution equals $P_{Y^{\pi_e}|X}=\sum_t \pi_e(t|X)P_{Y^t|X}$ if and only if $\pi_e$ is a deterministic policy or $\pi_b(0|X)=\pi_b(1|X)=\cdots=\pi_b(m-1|X)$.
\end{proposition}

\begin{figure*}
\centering
\includegraphics[width=0.85\linewidth]{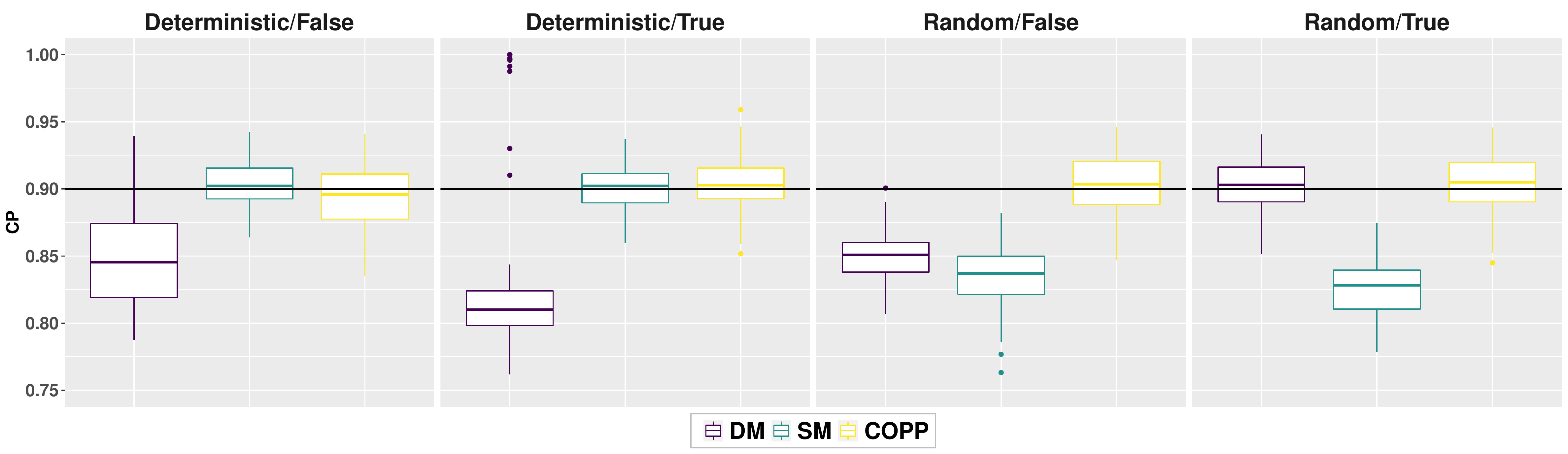}
\caption{Empirical coverage probabilities (CPs) of PIs based on the Direct method (DM), Subsampling-based method (SM) and our proposal (COPP) in single-stage studies. 
	The stochastic target policy is given by $\pi_e(1|X)=1-\pi_e(0|X)=\text{sigmoid}(-0.5+X^{(1)}+X^{(2)}-X^{(3)}-X^{(4)})$ and the deterministic target policy is given by $\mathbb{I}(X^{(3)}+X^{(4)}>X^{(1)}+X^{(2)})$. The nominal level $1-\alpha$ is $90\%$.}
\label{SS}
\end{figure*}

\textbf{Our proposal}. The subsampling-based method fails because the distribution of the selected response differs from that of the potential outcome. To address this issue, instead of sampling according to the target policy $\pi_e$, we carefully design a pseudo/auxiliary policy $\pi_a$ whose distribution depends on both $\pi_e$ and $\pi_b$ such that the resulting subsamples' distribution matches that of the potential outcome. More specifically, for any $0\le t<m-1$ and $x$, $\pi_a$ shall satisfy the following, 
\begin{eqnarray}\label{eqn:ratio}
\frac{\pi_a(t|x)}{\pi_a(0|x)}=\frac{\pi_e(t|x)}{\pi_e(0|x)}\left[ \frac{\pi_b(t|x)}{\pi_b(0|x)} \right]^{-1}.
\end{eqnarray}
In other words, $\pi_a(\bullet|x)$ shall be proportional to the ratio $\pi_e(\bullet|x)/\pi_b(\bullet|x)$ for any $x$. Similar to Proposition \ref{prop:1}, let $A$ denote the pseudo action generated according to $\pi_a$, we can show that subsamples with $A=T$ follow the following distribution,
\begin{eqnarray*}
P_{Y|A=T,X}=\sum_{t=0}^{m-1} \frac{\pi_{a}(t|X)\pi_b(t|X)}{\sum_{t'}\pi_{a}(t'|X)\pi_b(t'|X) }P_{Y^t|X}\\=\sum_{t=0}^{m-1} \pi_e(t|X) P_{Y^t|X}=P_{Y^{\pi_e}|X}.
\end{eqnarray*}  
This implies that subsampling according to the pseudo policy $\pi_a$ yields the same conditional distribution as $P_{Y^{\pi_e}|X}$ in the target population. Nonetheless, the selected subsamples and the target possess different covariate distributions. Such a ``covariate shift" problem can be naturally handled by the weighted CP algorithm. Using Lemma \ref{lemma1} again, the subsamples and the target population are weighted exchangeable with weights $w_i=1$ for any $i$ such that $A_i=T_i$ and
\begin{eqnarray*}
w_{n+1}(x,y)=\frac{P_{X,Y^{\pi_e}}(x,y)}{P_{X,Y|A=T}(x,y)}=\frac{P_X(x)}{P_{X|A=T}(x)}\\=\frac{\mathbb{P}(A=T)}{\mathbb{P}(A=T|X=x)}\propto\frac{1}{\mathbb{P}(A=T|X=x)},
\end{eqnarray*}
Compared to the direct method (see \ref{eqn:weight}), the weight in the above expression depends only on the behavior policy which is known in randomized studies. and is independent of $y$. Consequently, our proposal is robust to the model misspecification of the conditional distribution $P_{Y^t|X}$, as shown later. When the behavior policy is unknown, it can be estimated based on existing supervised learning algorithms. 
We summarize our proposal in Algorithm \ref{alg1}, and call our method COPP, short for conformal off-policy prediction. Finally, we remark that by \eqref{eqn:ratio}, $\pi_e=\pi_a$ only when $\pi_e$ is deterministic or $\pi_b$ is uniformly random. Consequently, the subsampling-based method is valid in these two special cases. 
\textbf{A numerical example}. We conduct a simulation study to further demonstrate the sub-optimality of the direct and subsampling-based methods. We generate 500 data points from Example 1 of Section \ref{sec:5} for calibration and 10000 test data points. 
We consider a random target policy and a deterministic target policy. We further consider two conditional distribution models for $Y^t|X$, corresponding to a correctly specified model (denoted by ``true"), and a misspecified model (denoted by ``false") generated by injecting uniformly random noises on $(0,1)$ to the oracle distribution function. It can be seen from Figure \ref{SS} that the direct method fails to cover the response when the conditional distribution model is misspecified whereas the subsampling-based method fails when the target policy is random. 
To the contrary, our proposal achieves valid coverage in all settings.

\begin{algorithm}[h]
\caption{COPP for single-stage decision making}\label{alg1}
\begin{algorithmic}[t]
	\State \textbf{Input:}
	Data $\{(X_i,T_i,Y_i)\}_{i=1}^n$; a test point $X_{n+1}$; a target policy $\pi_e$; number of treatment options $m$; propensity score training algorithm $\mathcal{P}$; quantile prediction algorithm $\mathcal{Q}$;  
	quantile levels $\alpha_U$, $\alpha_L$ with $\alpha_{U}-\alpha_{L}=1-\alpha$.
	\begin{itemize}
		\item[1:] Split the data into two disjoint subsets $\mathcal{Z}^{tr}\cup\mathcal{Z}^{cal}$.
		\item[2:] Estimate $\pi_b(t|x)$ via $\mathcal{P}$ using samples from $\mathcal{Z}^{tr}$. 
		\item[3:] Draw $\{A_i\}_{i=1}^n$ 
		by plugging the propensity score estimator $\widehat{\pi}_{b}(t|x)$ in \eqref{eqn:ratio}.
		\item[4:] Select subsamples satisfying $A_i=T_i$ in both data subsets. Denote them by $\mathcal{Z}^{tr,s}$ and $\mathcal{Z}^{cal,s}$. 
		\item[5:] Apply quantile regressions via $\mathcal{Q}$ to selected subsamples from $\mathcal{Z}^{tr,s}$ to obtain the conditional quantile functions $\widehat{q}_{\alpha_{L}}$ and $\widehat{q}_{\alpha_{U}}$.
		\item[6:] Compute the conformity scores $\{S_i\}_i$ for all selected subsamples $i\in \mathcal{Z}^{cal,s}$ according to \eqref{eqn:score}. 
		\item[7:] For any $i\in \mathcal{Z}^{cal,s}\cup\{n+1\}$, estimate the weight $\widehat{w}_{n+1}(X_i)=\textstyle\sum_{t=0}^{m-1}\pi_e(t|X_i)/\widehat{\pi}_{b}(t|X_i)$. 
		\item[8:] For any $i$, compute the mixing probability 
			$p_i^w=
				[\sum_{j\in\mathcal{Z}^{cal,s}\cup \{n+1\}} \widehat{w}_{n+1}(X_j)]^{-1}\widehat{w}_{n+1}(X_i)$.
			\item[9:] Compute $Q_{1-\alpha}(X_{n+1})$ as the $(1-\alpha)$th quantile of $\sum_{i\in\mathcal{Z}^{cal,s}}p_i^w \delta_{S_{i}}+p_{n+1}^w\delta_{\infty}$.
		\end{itemize}
		\State \textbf{Output:} the PI $\widehat{C}(X_{n+1})=[\widehat{q}_{\alpha_L}(X_{n+1})-Q_{1-\alpha}(X_{n+1}),\widehat{q}_{\alpha_U}(X_{n+1})+Q_{1-\alpha}(X_{n+1})]$
	\end{algorithmic}
\end{algorithm}

\textbf{Statistical properties}. Let $\widehat{\pi}_b(t|x)$ denote the estimated behavior policy, $w_{n+1}(x)=1/\mathbb{P}(A=T|X=x)$ denote the oracle normalized weight function and $\widehat{w}_{n+1}$ denote the estimated weight function in Step 7 of Algorithm \ref{alg1}. 
We first show that COPP achieves valid coverage \textit{asymptotically} when the behavior policy is consistently estimated. Notice that we do not require consistency of the estimated conditional outcome distribution. 

\begin{theorem}[Asymptotic coverage]\label{thm:AOP}
	Let $n_{1}=|\mathcal{Z}^{tr}|$. 
	Further, suppose that $\mathbb{E}[\widehat{w}_{n+1}(X)|\mathcal{Z}^{tr}]<\infty$, $\mathbb{E}[w_{n+1}(X)]<\infty$ and the consistency of behavior policy estimates (see the detailed requirements in Appendix A), then the output $\widehat{C}(x)$ from Algorithm \ref{alg1}  satisfies $$	\lim_{n_1\rightarrow\infty}\mathbb{P}
	(Y_{n+1}^{\pi_e}\in\widehat{C}(X_{n+1}))\ge 1-\alpha.$$	 
\end{theorem}

Next, we show that if the propensity scores are known in advance, the proposed PI achieves exact coverage in finite samples.

\begin{theorem}[Exact coverage]
	Suppose that  $\mathbb{E}[w_{n+1}(X)]<\infty$, then the output $\widehat{C}(x)$ from Algorithm \ref{alg1} with correctly specified propensity scores satisfies, for any sample size $n$,
	\begin{equation*}
		\mathbb{P}
		(Y_{n+1}^{\pi_e}\in\widehat{C}(X_{n+1}))\ge 1-\alpha.
	\end{equation*}
	
\end{theorem}

Finally, we show that the proposed PI is asymptotically efficient when the quantile regression estimator in Step 5 of Algorithm \ref{alg1} is consistent. 
\begin{theorem}[Asymptotic efficiency]
	Suppose the behavior policy is known and 
	the quantile regresssion estimates are consistent (see the detailed requirements in Appendix A), the output $\widehat{C}(x)$ from Algorihtm \ref{alg1} satisfies
	\begin{equation*}
		L(\widehat{C}(X_{n+1})\triangle C_{\alpha}^{\text{oracle}}(X_{n+1}))=o_{p}(1),
	\end{equation*}
	as $|\mathcal{Z}^{tr}|,|\mathcal{Z}^{cal}|\rightarrow\infty$. Here $L(A)$ indicates the Lebesgue measure of the set $A$, and $\triangle$ is the symmetric difference operator, i.e., $A\triangle B=(A \backslash B)\cup(B\backslash A)$, $C_{\alpha}^{\text{oracle}}(x)$ is the oracle interval defined as $[q_{\alpha_{L}}(x),q_{\alpha_{U}}(x)].$
\end{theorem}

\subsection{Extensions}\label{sec:ext}
In this section, we discuss two extensions of COPP, based on importance sampling and multi-sampling, respectively. 

\textbf{Extension 1}. One limitation of COPP lies in that the PIs are constructed based only on observations in the subsamples. Nonetheless, when the target policy is stochastic, each observation has certain chance of being selected. To make full use of data, we adopt the importance sampling trick 
to compute the normalized weights and quantiles in Steps 8 and 9 of Algorithm \ref{alg1}, respectively. Specifically, in Step 7, we set the weight $\widehat{w}_{n+1}(X_i)$ for each of the sample in $\mathcal{Z}^{cal}$ to $\widehat{\pi}_a(T_i|X_i)\widehat{w}_{n+1}(X_i)$. 
These weights are then passed to Step 8 to compute $\widehat{p}_i$, and subsequently to Step 9 to calculate $Q_{1-\alpha}(X_{n+1})$ by replacing $\mathcal{Z}^{cal,s}$ with the whole calibration set $\mathcal{Z}^{cal}$. As we will show in Section \ref{sec:5}, this procedure is much more efficient than COPP when the selected subsamples contains only a few observations. We next prove that such an extension achieves valid coverage as well. 

\begin{theorem}\label{thm:AEOP}
	Under the conditions of Theorem \ref{thm:AOP}, we have 
	\begin{equation*}
		\lim_{n_1\rightarrow\infty}\mathbb{P}
		(Y_{n+1}^{\pi_e}\in\widehat{C}(X_{n+1}))\ge 1-\alpha.
	\end{equation*}
\end{theorem}


\textbf{Extension 2}. The second extension integrates COPP with the multi-sampling method. Notice that Algorithm \ref{alg1} only implements subsampling once. The result can be very sensitive to the selected subsamples.  
To mitigate the randomness the single-sampling procedure introduces, we propose to repeat COPP 
multiple times and then aggregate all these PIs to gain efficiency. 
To combine multiple PIs, we adopt the idea proposed by \cite{solari2022multi} 
for multi-split conformal prediction. 
A key observation is that, the PI in Algorithm \ref{alg1} is equivalent to $\widehat{C}(X_{n+1})=\{y: p(X_{n+1},y)\ge\alpha\}$
where $p(X_{n+1},y)$ is given by $$\sum_{i\in \mathcal{Z}^{cal,s}}p_i^w\mathbb{I}[\max\{\widehat{q}_{\alpha_{L}}(X_{n+1})-y,y-\widehat{q}_{\alpha_{U}}(X_{n+1})\}\le S_i]+p_{\infty}^w,$$ 
serving as a $p$-value for the testing hypotheses  $H_{0}:Y_{n+1}^{\pi_e}=y$ against $H_{1}:Y_{n+1}^{\pi_e}\ne y$ given $X_{n+1}$. This allows us to follow the idea of \citet{meinshausen2009p} for $p$-value aggregation. 
Let $p^{b}(x,y)$ for $1\le b\le B$ be the $p$-values for $B$ constructed PIs with significance level $\alpha\gamma$ for certain  tuning parameter $0<\gamma<1$. We aggregate these $p$-values by setting $\bar{p}(X_{n+1},y)$ to their empirical $\gamma$-quantile. The final PI is given by $\widehat{C}_{B,\gamma}(X_{n+1})=\{y: \bar{p}(X_{n+1},y)\ge \alpha\}$. 
\begin{theorem}\label{thm:boot}
	Under the conditions of Theorem \ref{thm:AOP}, we have for any $B>0$ and $0<\gamma<1$,
	\begin{equation*}
		\lim_{n_1 \rightarrow\infty}\mathbb{P}
		(Y_{n+1}^{\pi_e}\in\widehat{C}_{B,\gamma}(X_{n+1}))\ge 1-\alpha.
	\end{equation*}
\end{theorem}

Finally, we remark that we only derive the asymptotic coverage of the two extensions in Theorems \ref{thm:AEOP} and \ref{thm:boot}. Nonetheless, when the behavior policy is known, these methods also achieve exact coverage. 

\section{Conformal Off-Policy Prediction in Sequential Decision Making}\label{sec:4}

\textbf{Problem formulation}. In this section, we consider sequential desicion making where the observed data consist of $n$ i.i.d samples $\{(X_{1i},T_{1i},X_{2i},T_{2i},\ldots,X_{Ki},T_{Ki},Y_{i})\}_{i=1}^{n}$ where for the $i$th instance, $X_{ki}$ collects the state information at the $k$th stage, $T_{ki}\in\{0,\ldots,m-1\}$ denotes the action at the $k$th stage , $Y_{i}$ is the corresponding reward at the final stage. Such a sparse reward setting is frequently considered for precision medicine type applications \citep{murphy2003optimal}. Meanwhile, our method is equally applicable to settings with immediate rewards at each decision point (see Appendix B). 

Let $H_{k}=\{X_1,T_1,\ldots,X_k\}$ denote the history up to the $k$th stage. 
We define a (history-dependent) policy $\Pi=(\pi_{1}(t_1|h_1),\pi_{2}(t_2|h_2),\ldots,\pi_{K}(t_K|h_K))$ as a sequence of (stochastic) decision rules where each $\pi_{k}(t_k|h_k)$ determines the probability that an agent selects action $t_k$ at the $k$th stage given that $H_{k}=h_k$. For a given target policy $\pi_e$, we are interested in constructing PIs for the potential outcome $Y^{\pi_e}$ that would be observed were the instance to follow $\pi_e$ for any initial state $X_1$. To save space, we impose the consistency, sequential ignorability and positivity assumption in Appendix B. 


\textbf{COPP}. We generalize our proposal in Section \ref{sec:challenge} to
sequential making decision. We design a pseudo policy $\pi_{a}=\{\pi_{a,k}\}_k$ which relies on both $\pi_b=\{\pi_{b,k}\}_k$ and $\pi_e=\{\pi_{e,k}\}_k$, to generate subsamples whose outcome distribution conditional on the state-action history matches that of the potential outcome. Specifically, for any $1\le k\le K$ and $h_k$, the pseudo policy $\pi_{a,k}(\bullet|h_k)$ shall be proportional to the ratio $\pi_{e,k}(\bullet|h_k)/{\pi_{b,k}(\bullet|h_k)}$. 
Similar to Proposition \ref{prop:1}, 
we can show that the conditional density of $Y|A_K=T_K,H_{K}$ equals that of $Y^{\pi_e}|H_{K}$.

More importantly, by iteratively integrating over the space of $\{T_k,X_{k+1},\cdots,X_K\}$, we can show that
the conditional density of $Y|A_k=T_k, \cdots, A_K=T_K,H_k$ also equals that of $Y^{\pi_e}|H_k$ for each $k$. Using Lemma \ref{lemma1} again, the subsamples $\{(H_{1i},Y_{1i}): A_{ki}=T_{ki}, 1\le k\le K, 1\le i\le n \}$ and the target population $(H_{1,n+1},Y_{n+1})$ are weighted exchangeable with weights $w_i=1$ for any $i$ and 
\begin{equation*}
	w_{n+1}(h)\propto \mathbb{P}^{-1}(A_1=T_1,\cdots,A_K=T_K|H_1=h).
\end{equation*}
Based on these weights, the PIs can be similarly derived as in Algorithm \ref{alg1}. We defer
the pseudocode and the statistical properties of the constructed PIs to Appendix B.

\textbf{Extensions}. Our proposal suffers from the ``curse of horizon" \citep{liu2018breaking} in that the number of selected subsamples decreases exponentially fast with respect to the number of decision stages. While this phenomenon is unavoidable without further model assumptions \citep{jiang2016doubly}, the importance-sampling-based and multi-sampling-based approach alleviate this issue to some extent, as shown in our simulations. Since these extensions are very similar to those presented in Section \ref{sec:ext}, we omit them for brevity. 

\section{Synthetic Data Analysis}\label{sec:5}
In this section, we conduct simulation studies to investigate the empirical performance of our proposed methods. 
In particular, we focus on the following three examples: two examples considered in  \citet{wang2018quantile}:

\textbf{Example 1 (Single-Stage Decision Making):}
\begin{itemize}[leftmargin=*]\vspace{-0.2cm}
	\item The baseline covariates $X^{(1)},X^{(2)},X^{(3)},X^{(4)}$ are independently uniformly generated from $(0,1)$. \vspace{-0.2cm}
	\item The action is binary and 
	satisfies $\mathbb{P}(T=1|X)= \mbox{sigmoid}(-0.5-0.5\sum_{j=1}^4 X^{(j)})$ where $\mbox{sigmoid}(t)=\exp(t)/[1+\exp(t)]$. \vspace{-0.2cm}
	\item The return is given by $Y=1+X^{(1)}-X^{(2)}+(X^{(3)})^3+\exp(X^{(4)})+T(3-5X^{(1)}+2X^{(2)}-3X^{(3)}+X^{(4)})+(1+T)(1+\sum_{j=1}^4 X^{(j)})\epsilon$ where $\epsilon$ is a standard normal variable independent of $X$ and $T$. \vspace{-0.2cm}
	\item  The target policy $\pi_e$ satisfies $\pi_e(1|X)=\mbox{sigmoid}(-0.5+X^{(1)}+X^{(2)}-X^{(3)}-X^{(4)}))$.
\end{itemize}

\textbf{Example 2 (Two-Stage Decision Making):} 
\begin{itemize}[leftmargin=*]\vspace{-0.2cm}
	\item Observations and actions are generated as follows:
	\begin{eqnarray*}
		X_1\sim \textrm{Uniform}(0,1),\\ T_1|X_1\sim \textrm{Bernoulli}(\textrm{sigmoid}(-0.5+X_1)),\\
		X_2|X_1,T_1\sim \textrm{Uniform}(X_1,X_1+1),\\ T_2|X_1,T_1,X_2\sim\mbox{Bernoulli}(\mbox{sigmoid}(-0.5-X_2)).
	\end{eqnarray*}
	\item The final return is given by
	$Y=1+X_1+T_1[1-3(X_1-0.2)^2]+X_2+T_2[1-5(X_2-0.4)^{2}]+(1+0.5T_1-T_1X_1+0.5T_2-T_2X_2)\epsilon$ for a standard normal variable $\epsilon$ independent of observations and actions.  
	\item The target policy is defined as follows
	\begin{eqnarray*}
		E_1|X_1\sim\mbox{Bernoulli}(\mbox{sigmoid}(0.5X_1-0.5)),\\ E_2|X_1,E_1,X_2\sim\mbox{Bernoulli}(\mbox{sigmoid}(0.5X_2-1)).
	\end{eqnarray*} 
\end{itemize}
For each example, we further consider two settings. In the high-dimensional setting, we manually include $100-p_0$ null variables that are uniformly distributed on $(0,1)$ in the state with $p_0=4$ and $1$ in Examples 1 and 2, respectively. In the low-dimensional setting, these null variables are not included.
This yields a total of four different scenarios. The sample size is fixed to $2000$.

\textbf{Example 3 (Multi-Stage Decision Making):} We design an additional simulation studies where the number of horizon (e.g., the decision stages) equals 3, 4 or 5, and investigate the performance of our methods under this setting.
\begin{itemize}[leftmargin=*]\vspace{-0.2cm}
	\item Observations and actions are generated as follows:
	\begin{eqnarray*}
		X_1=0.5\epsilon_{1},\epsilon_{k}\sim N(0,1), 1\le k\le  m,\\ T_{k}|X_{k}\sim\mbox{Bernoulli}(\mbox{sigmoid}(-0.5+X_k)),\\
		X_{k}=0.5X_{k-1}+0.1T_{k-1}+0.5\epsilon_{k}.
	\end{eqnarray*}
	\item The final return is given by
	$Y_m=X_m$.
	\item The target policy is defined as follows
	\begin{eqnarray*}
		D_{k}|X_k\sim\mbox{Bernoulli}(\mbox{sigmoid}(-0.5+0.5X_k)).
	\end{eqnarray*} 
\end{itemize}
The data consist of 2000 observations, in which three quarters are used for training and the rest for validation. 

\textbf{Implementation details}. We estimate the behavior policy using logistic regression. In the high-dimensional setting, we apply penalized regression to improve the estimation efficiency. The conditional quantile functions are estimated based on quantile regression forest \citep{meinshausen2006quantile}. Following \citet{sesia2020comparison}, we use 75\% of the data for training and the rest for calibration. We fix $\alpha_{L}=\alpha/2$ and $\alpha_{U}=1-\alpha/2$ in all settings. In addition, to implement the multi-sampling-based method, we fix $\gamma=1/2$ and set the significance level to $\alpha$ instead of $\alpha\gamma=\alpha/2$ to improve the precision (interval length). We find that the resulting PI achieves nominal coverage in practice.    
The number of intervals $B$ is set to 100 in the low-dimensional setting, and $50$ in high dimension to reduce computation time. 
Finally, in each simulation, we generate $10000$ test data points in the target population to evaluate the converge probability. The R code is released \href{https://github.com/yyzhangecnu/COPP}{here}. 

\textbf{Benchmark specification}. We compare our proposed methods against the subsampling-based method (SM) detailed in Section \ref{sec:challenge}. In low-dimensional settings, we also compare with the standard importance sampling (IS) and doubly robust (DR) method \citep[see e.g.,][]{dudik2011doubly,zhang2012robust,jiang2016doubly} designed for off-policy confidence interval estimation. These methods focus on the average effect. We couple them with kernel density estimation to infer the individual effect conditional on the initial state. Please refer to Appendix C for the detailed implementation.
\begin{figure*}
	\centering
	\subfloat{\includegraphics[width=0.22\linewidth]{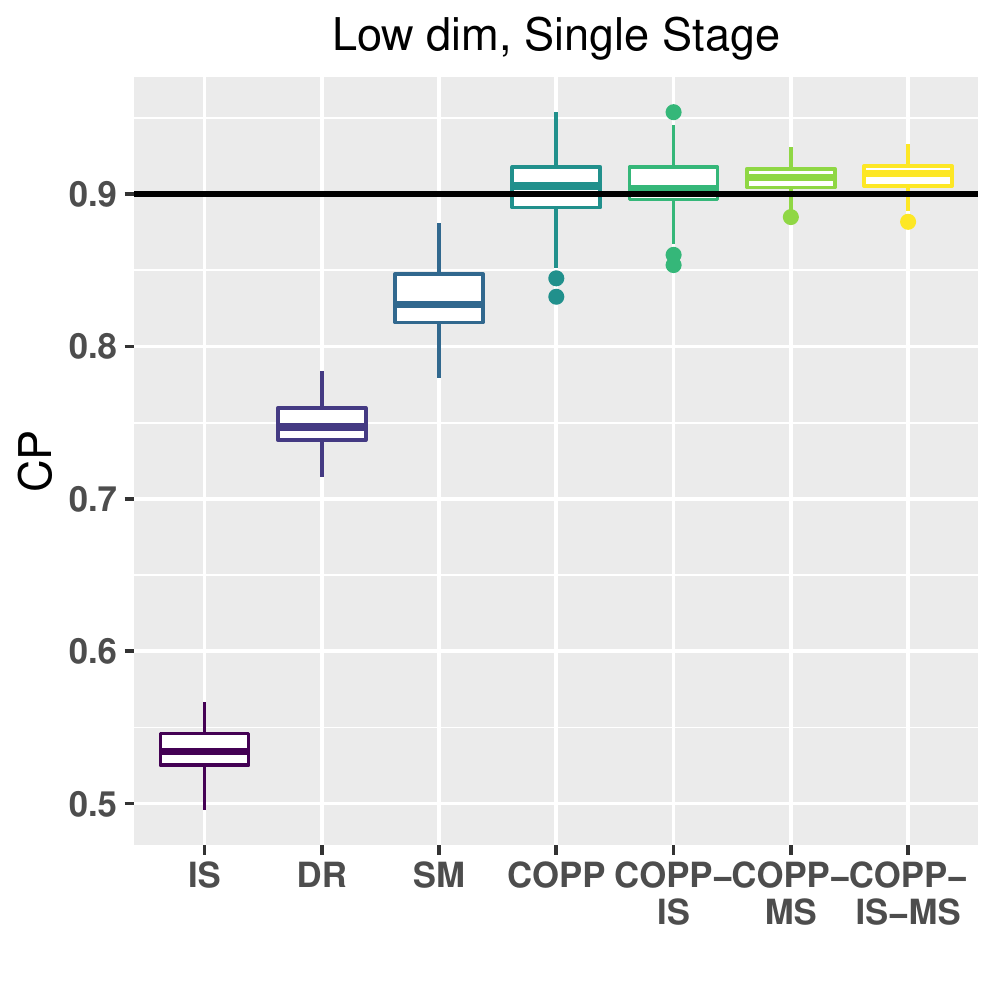}}
	\subfloat{\includegraphics[width=0.22\linewidth]{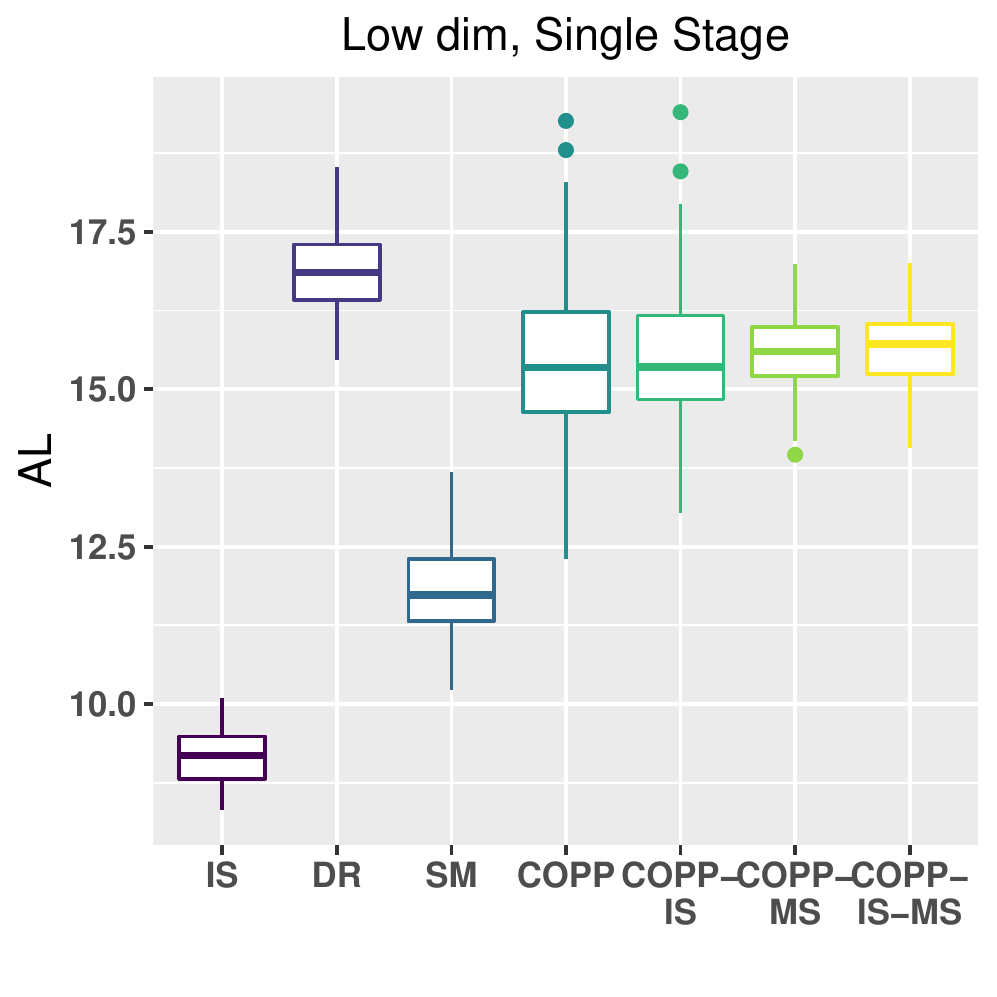}}
	\subfloat{\includegraphics[width=0.22\linewidth]{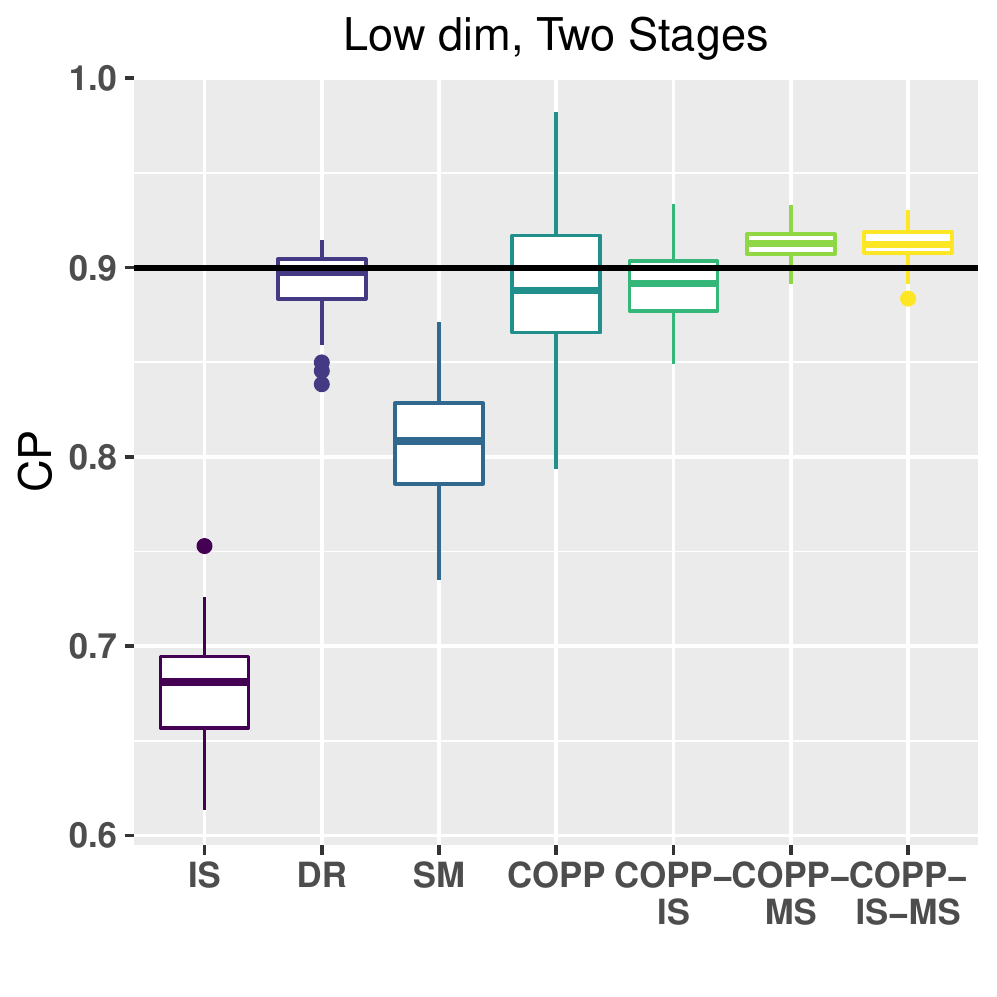}}
	\subfloat{\includegraphics[width=0.22\linewidth]{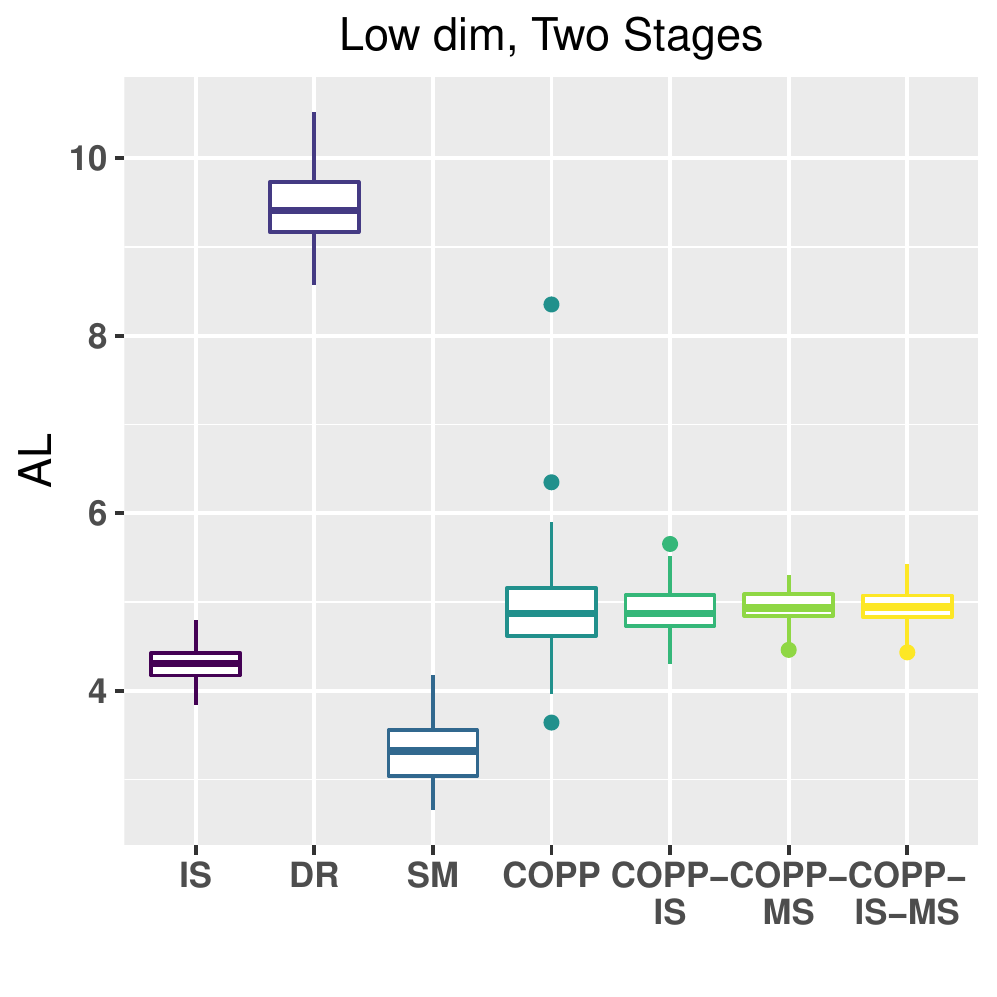}}\\
	\subfloat{\includegraphics[width=0.22\linewidth]{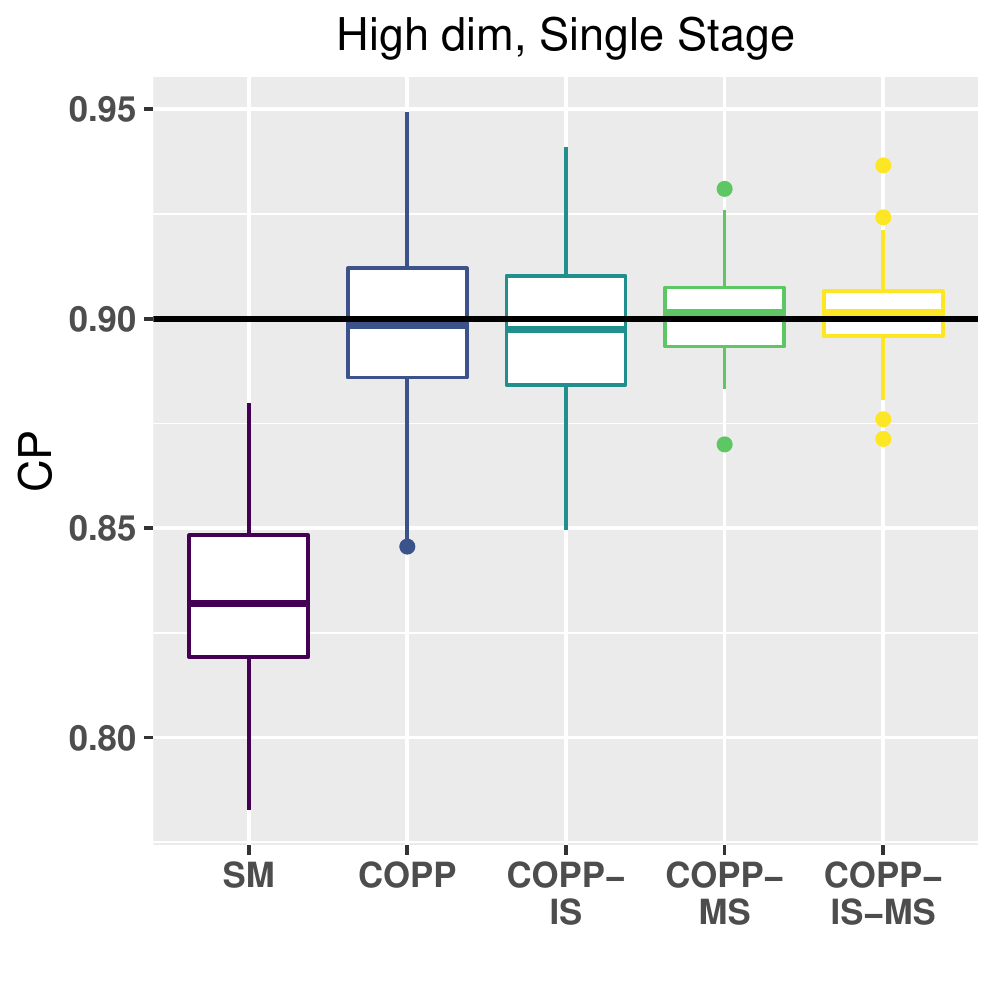}}
	\subfloat{\includegraphics[width=0.22\linewidth]{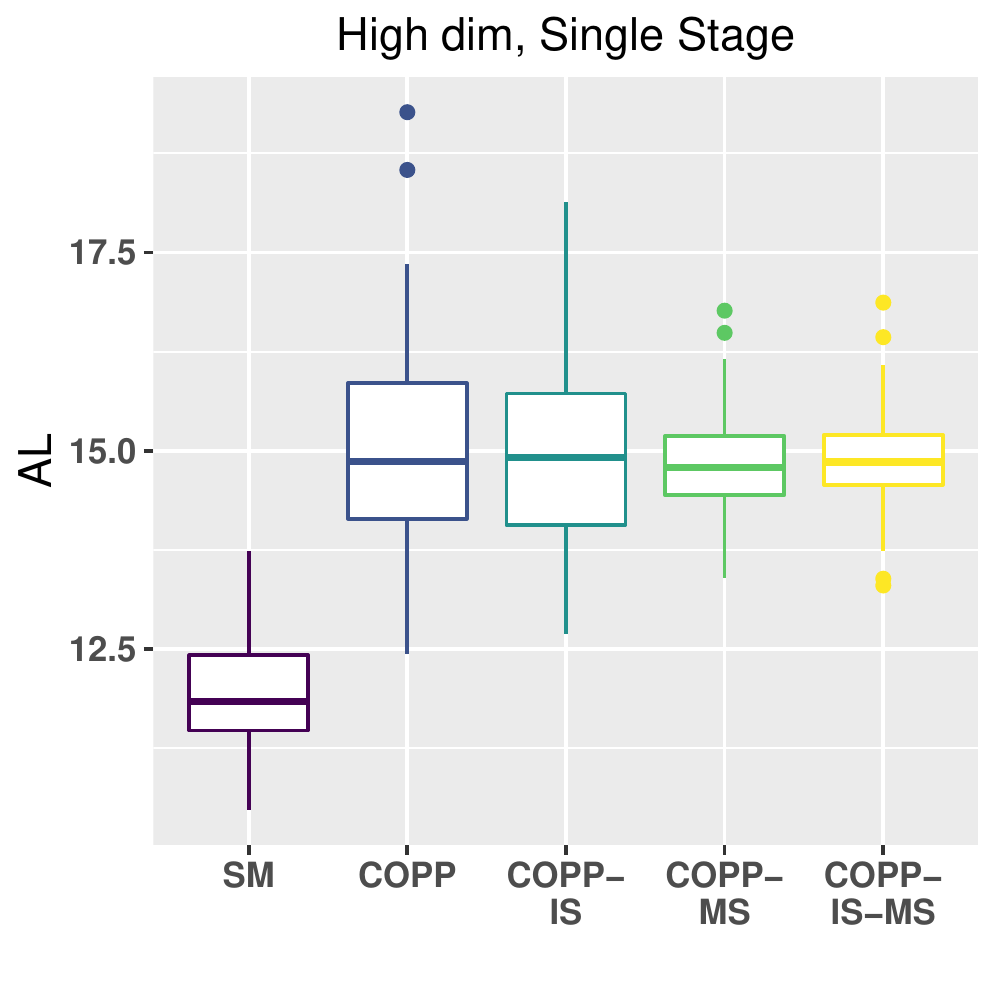}}
	\subfloat{\includegraphics[width=0.22\linewidth]{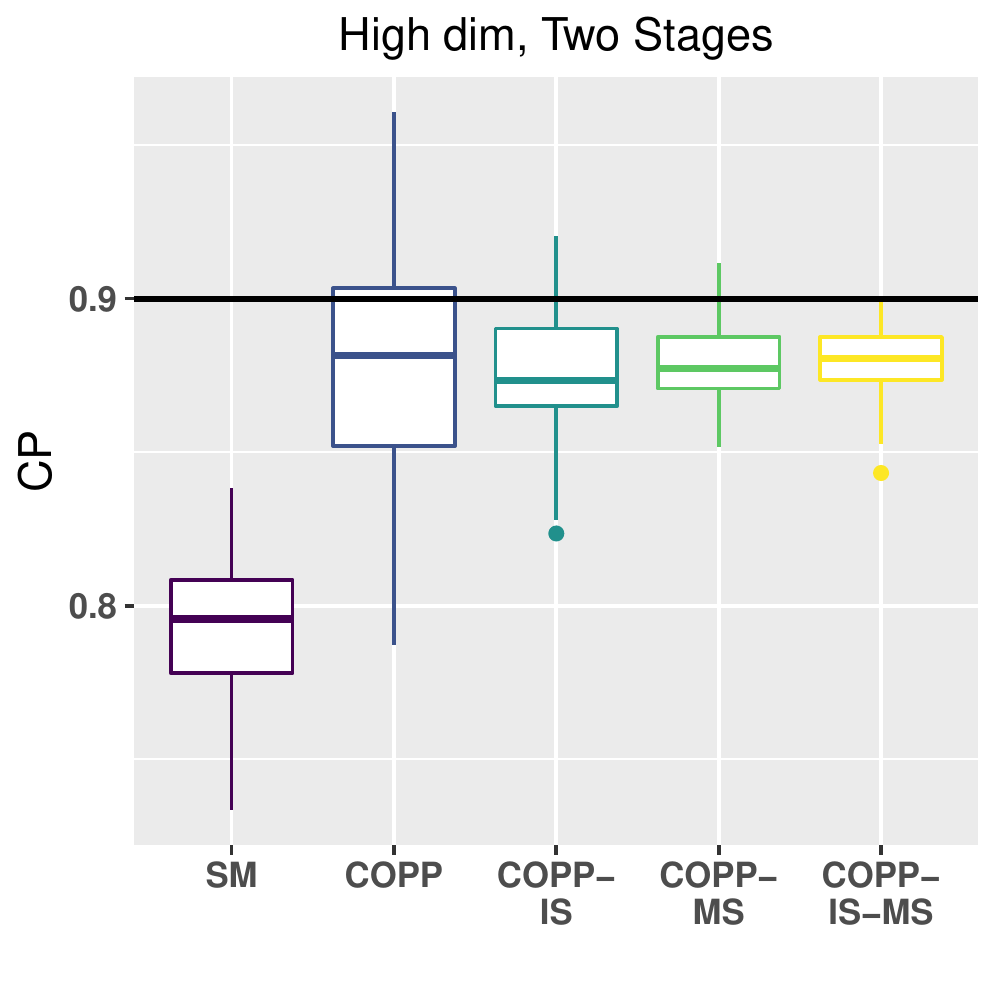}}
	\subfloat{\includegraphics[width=0.22\linewidth]{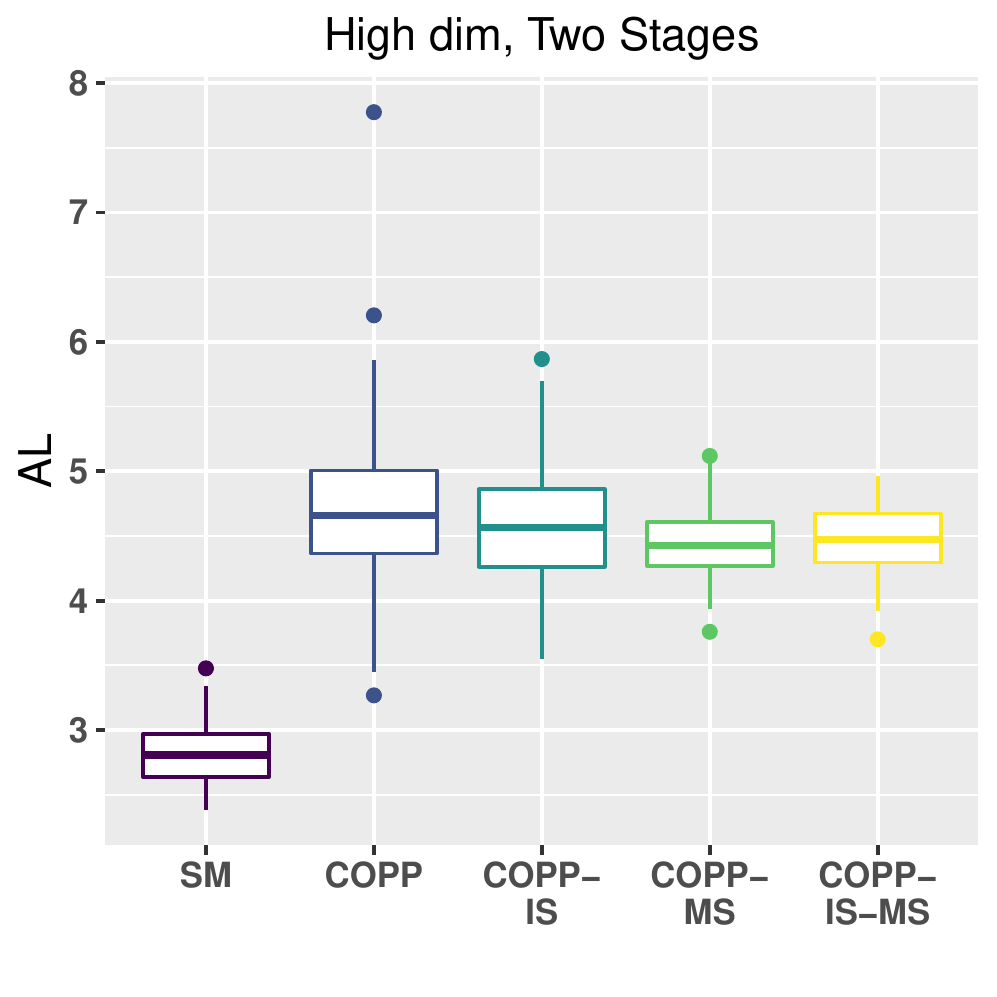}}
	\caption{Empirical coverage probabilities (CPs) and average lengths (ALs) of intervals based on SM, IS, DR, and our proposed COPP, COPP-IS, COPP-MS, COPP-IS-MS in four settings. The nominal level is $90\%$.}
	\label{Low-dim}
\end{figure*}

\textbf{Results}. Figure \ref{Low-dim}
reports the coverage probability and average length of various interval estimators for Examples 1 and 2, aggregated over 100 simulations. We denote the extensions of our proposal based on importance-sampling, multi-sampling alone and a combination of the two are denoted by COPP-IS, COPP-MS and COPP-IS-MS, respectively. We summarize our findings below. First, intervals based on SM, IS and DR significantly undercover the potential outcome. As we have commented, these methods are not valid in general. SM requires either a uniformly random behavior policy, or a deterministic target policy. IS and DR focus on the expected return and ignore the variability of the return around its expectation. 
Second, all the proposed methods achieve nominal coverage in most cases. Among them,
the multi-sampling-based methods (COPP-MS and COPP-IS-MS) achieve the best performance, with substantially reduced variability compared to the single-sampling-based methods. In addition, COPP-IS performs much better than COPP in two-stage settings where the number of subsamples is limited, as expected. 

\begin{figure*}
	\begin{minipage}[b]{0.33\linewidth}
		\includegraphics[width=\linewidth]{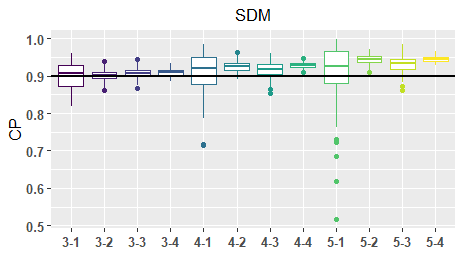}
		\caption{Coverage Probability for SDM for Horizons 3,4,5.}\label{SDM}
	\end{minipage}
	\begin{minipage}[b]{0.33\linewidth}
		\includegraphics[width=\linewidth]{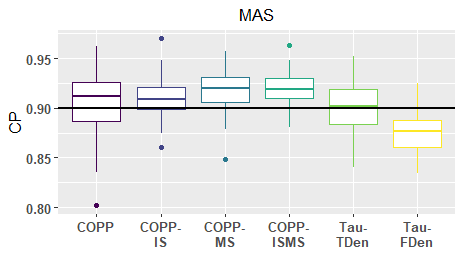}
		\caption{Coverage Probability for Multiple Action Space}\label{MAS-CP}
	\end{minipage}
	\begin{minipage}[b]{0.33\linewidth}
		\includegraphics[width=\linewidth]{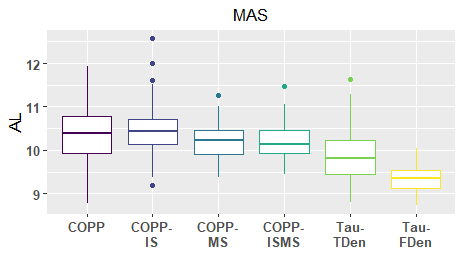}
		\caption{Average Length for Multiple Action Space}\label{MAS-AL}
	\end{minipage}
\end{figure*}

We report the results for Example 3 in Figure \ref{SDM}, where the proposed COPP, COPP-IS, COPP-MS, COPP-IS-MS with horizon $m$ are labelled as $m$-1, $m$-2, $m$-3 and $m$-4, respectively. It can be seen that the proposed method is able to achieve nominal coverage in general. Nonetheless, as commented in our paper, it suffers from the curse of horizon and would be inefficient in long-horizon settings. It remains unclear how to break the curse of horizon and we leave it as future work.

\textbf{Comparison with \citet{taufiq2022conformal}.} 
We consider the simulation setting described in Section 6.1.1 of \citet{taufiq2022conformal} with four actions, implement their methods with a correctly specified $f$ (denoted by ``Tau-TDen'') and a misspecified model (denoted by``Tau-FDen''), and compare both methods against our proposal with a misspecified $f$. All these methods additionally require to specify the propensity score model and we use a correctly specified model. It is seen from Figures \ref{MAS-CP} and \ref{MAS-AL} below that unlike our proposal, Taufig et al. (2022)'s method is sensitive to the specification of the conditional density function.

\section{Real Data Analysis}\label{sec:real}
\begin{figure}[t]
	\centering
	\includegraphics[width=\linewidth]{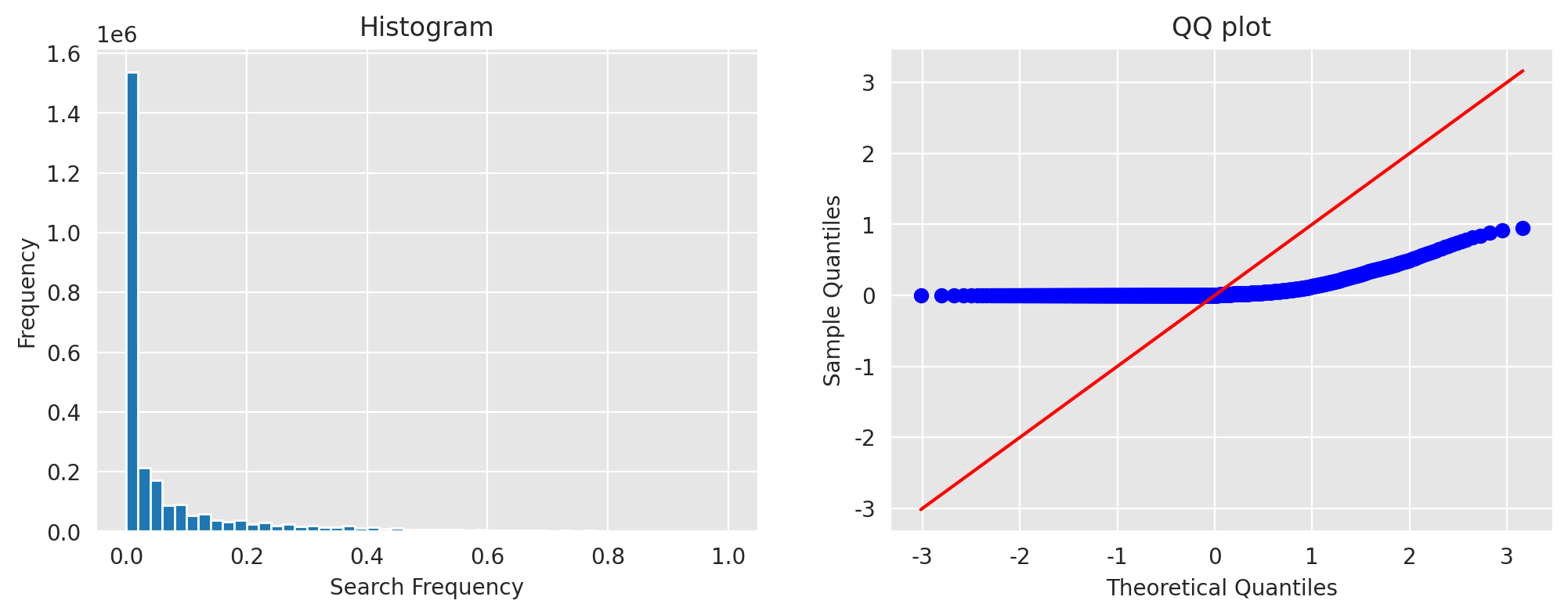}
	\caption{Histogram and QQ plot of users' search frequencies. In the QQ plot, the blue curve depicts the empirical quantiles of the search frequency against those from a standard normal distribution function. }
\label{LLP}
\end{figure}
We illustrate the usefulness of our method based on a dataset collected from a world-leading technology company. This company has one of the largest mobile platforms for production, aggregation and distribution of short-form videos with extensive search functionality. It implements a strategy to encourage users to explore its search functionality. Specifically, when a user launches the app for the first time in a day, they will see a pop-up window that recommends them to use the search feature. However, pop-ups are annoying for some users. As such, the company's interested in `pop-up' policies that implement this strategy to a subgroup of users to increase their search frequency. 

The dataset is collected from an online experiment which involves two millions daily active users and has been scaled due to privacy. The features available to us consist of each user's history information including the frequency they used the app and the search functionality prior to the experiment. The reward is the user's search frequency after treatment and is highly heavy-tailed as shown 
in Figure \ref{LLP}. As such, instead of focusing on a target policy's expected return, we are interested in its entire distribution. As commented earlier, most existing OPE methods are not directly applicable.  In addition, since the behavior policy is known to us, the proposed method is robust to the model misspecification of the outcome distribution, and achieves exact coverage. 

To investigate the validity of the proposed method, we equally split the dataset into two, one for learning an optimal policy and the other for policy evaluation. On the evaluation dataset, we further employ ten-fold cross-validation to test our method. Specifically, 
we randomly split the data into ten folds, use nine of them to train the proposed PIs and the remaining fold to estimate their coverage probabilities. We further aggregate these coverage probabilities over different folds to get the full efficiency. 
In our implementation, the number of intervals $B$ is set to 100, and other configurations are consistent with those in Section \ref{sec:5}.  It turns out that the average coverage probabilities of COPP, COPP-MS, COPP-IS and COPP-IS-MS are all close to the nominal level 90\%, and equal 91.0\%, 90.5\%, 89.6\% and 90.4\%, respectively. The lower and upper bounds of the proposed PIs offer a more accurate characterization of the target policy's return and give practitioners more information when conducting online A/B tests.

\section{Conclusion and Discussion}
To our knowledge, our proposal is the first to procedure statistically sound PIs for a target policy's return in sequential decision making. The proposed PIs focus on the individual effect, take the variability of the return around its mean into consideration, achieve finite-sample coverage guarantees and are robust to the misspecification of the conditional outcome model. Currently, we consider a discrete action space. It would be practically interesting to extend our proposal to the continuous action setting. However, this is beyond the scope of the current paper and we leave it for future research.

\section*{Acknowledgements}
Zhang's research is supported by the National Natural Science Foundation of China (NSFC) grant nos. 12101241. Shi's research is partly support by the EPSRC grant EP/W014971/1. The authors wish to thank all the constructive comments from the referees, the area chair and the chair, which has lead to a significant improvement of the earlier version
of this paper.

\bibliography{CPOPE}
\bibliographystyle{apalike}

\appendix
\onecolumn

\section{PROOF OF MAIN RESULT}
This section provides the additional assumptions and proofs for Theorems 1-5 for COPP, COPP-IS, COPP-MS in contextual bandits. We first provide two assumptions concerning the consistency of behavior policy estimates $\widehat{\pi}_b(t|x)$. Assumption 1 is similar to A1 in 
\cite{lei2021conformal}. Let $\widetilde{A}$ and $A$ 
denote the pseudo actions generated according to the estimated behavior policy $\widehat{\pi}_b$ obtained from the training dataset $\mathcal{Z}^{tr}$ and the oracle behavior policy $\pi_b$, respectively. Let ${P}_{X|\widetilde{A}=T}$ be the distribution function of the contextual information in selected samples $\mathcal{Z}^{cal, s}$ as an estimate of $P_{X|A=T}$. Since observations in the calibration dataset follow same distribution, it suffices to estimate the weight function $w_{n+1}$. To ease notation, we will omit the subscript $n+1$ in $w_{n+1}$ and $\widehat{w}_{n+1}$ when there is no confusion. 


\begin{assumption}\label{ass:ss1}
	For any $0\le t\le m-1$, $\lim\limits_{n_{1}\rightarrow\infty}\mathbb{E}_{X\sim P_{X}}\left|1/\widehat{\pi}_b(t|X)-1/\pi_b(t|X)\right|\rightarrow 0$.
\end{assumption}

\begin{assumption}\label{ass:ss2}
	$\mathbb{E}_{X\sim P_{X|\widetilde{A}=T}}[\widehat{w}(X)|\mathcal{Z}^{tr}]<\infty$. 
\end{assumption}
\begin{proof}[\textbf{Proof of Theorem 1 (Asymptotic coverage)}]
	The proof is similar to that of Theorem 3 in \cite{lei2021conformal}. First, we index the data points in the selected calibration dataset $\mathcal{Z}^{cal, s}$ by $\{1,2,\ldots,n_2^\prime\}$.
	Notice that ${P}_{X|\widetilde{A}=T}$ is close to ${P}_{X|A=T}$ given a consistent behavior policy estimator. In addition, the two conditional distributions are the same if the propensity score is known in advance. The distribution of selected outcome is given by ${P}_{Y|\widetilde{A}=T,X}$, which aims to approximate ${P}_{Y|{A}=T,X}$. By definition, we have
	\begin{eqnarray}
		{P}_{Y|\widetilde{A}=T,X}&=&\sum_{t=0}^{m-1}\frac{\pi_e(t|x)\pi_b(t|x)/\widehat{\pi}_b(t|x)}{\sum_{t^\prime}\pi_e(t^\prime|x)\pi_b(t^\prime|x)/\widehat{\pi}_b(t^\prime|x)}P_{Y^{t}|X}.\notag
	\end{eqnarray}
	
	Using Bayesian rule,
	\begin{eqnarray}
		&&d{P}_{X|\widetilde{A}=T}(x)=\frac{{P}(\widetilde{A}=T|X=x)dP_{X}(x)}{\int_{x}{P}(\widetilde{A}=T|X=x)dP_{X}(x)}\notag\\
		&&=\frac{(\sum_{t=0}^{m-1}\pi_e(t|x)\pi_b(t|x)/\widehat{\pi}_b(t|x))/(\sum_{t=0}^{m-1}\pi_e(t|x)/\widehat{\pi}_b(t|x))dP_{X}(x)}{\int_{x}(\sum_{t=0}^{m-1}\pi_e(t|x)\pi_b(t|x)/\widehat{\pi}_b(t|x))/(\sum_{t=0}^{m-1}\pi_e(t|x)/\widehat{\pi}_b(t|x))dP_{X}(x)}.\label{eqn:condx}
	\end{eqnarray}
	
	Since the denominator is a constant and the weighted conformal inference is invariant to rescaling of the likelihood ratio,  we can reset the estimated weights by multiplying the denominator in the above. In other words, instead of setting $\widehat{w}(x)$ to $\sum_{t=0}^{m-1}\pi_e(t|x)/\widehat{\pi}_b(t|x)$ as in the main paper, we define
	\begin{equation}
		\widehat{w}(x)=\frac{\int_{x}(\sum_{t=0}^{m-1}\pi_e(t|x)\pi_b(t|x)/\widehat{\pi}_b(t|x))/(\sum_{t=0}^{m-1}\pi_e(t|x)/\widehat{\pi}_b(t|x))dP_{X}(x)}{1/(\sum_{t=0}^{m-1}\pi_e(t|x)/\widehat{\pi}_b(t|x))}.\label{eqn:weight}
	\end{equation}
	
	%
	
	Let $Z_i=(X_i,Y_i)$ for $1\le i\le n_2^\prime$ where $(X_{i},Y_{i})\sim {P}_{X|\widetilde{A}=T}\times {P}_{Y|X,\widetilde{A}=T}$  and $Z_{n+1}=(X_{n+1},Y_{n+1})\sim P_{X}\times P_{Y^{\pi^e}|X}$. Recall that $P_{Y|X,A=T}=P_{Y^{\pi^e}|X}$ and $P_{Y|X,\widetilde{A}=T}=P_{Y^{\pi^e}|X}$ if and only if $\widehat{\pi}_b(t|x)=\pi_b(t|x)$ for any $0\le t \le m-1$ and $x$.  The true weight function $w$ is set to $$w(x)=\frac{dP_{X}(x)}{d{P}_{X|A=T}(x)},$$
	accordingly. Under the assumption that $\mathbb{E}[w(X)]<\infty$, we have $\mathbb{P}_{X\sim P_{X}}(w(X)<\infty)=1$.	Similarly, under the assumption that $\mathbb{E}[\widehat{w}(X)|\mathcal{Z}^{tr}]<\infty$, we obtain 
	%
	%
	\begin{equation*}
		\mathbb{P}_{X\sim P_{X}}(\widehat{w}(X)<\infty)=1.
	\end{equation*}	
	%
	Let $\widetilde{P}_{X}$ denote a context distribution function such that $d\widetilde{P}_{X}(x)$ is proportional to $\widehat{w}(x)d{P}_{X|\widetilde{A}=T}(x)$. 
	Under Assumption \ref{ass:ss2}, we have that $\mathbb{P}_{X\sim {P}_{X|\widetilde{A}=T}}(\widehat{w}(X)<\infty|\mathcal{Z}^{tr})=1$, which in turn implies that $\widetilde{P}_{X}$ is a probability measure. Consider now a new sample $\widetilde{Z}_{n+1}=(\widetilde{X}_{n+1},\widetilde{Y}_{n+1})\sim \widetilde{P}_{X}\times{P}_{Y|\widetilde{A}=T,X}$.
	
	Let $E_{\widetilde{z}}$ denote the event that $\{Z_1,\ldots,Z_{n_2^\prime},\widetilde{Z}_{n+1}\}=\{z_1,\ldots,z_{n_2^\prime},\widetilde{z}_{n+1}\}$. The corresponding nonconformity scores are denoted as $\widetilde{S}=(S_1,\ldots,S_{n_2^\prime},\widetilde{S}_{n+1})$, $s_{i}=\mathcal{S}(z_i,\mathcal{Z}^{tr})$ for $1\le i\le n_2^\prime$ and $\widetilde{s}_{n+1}=\mathcal{S}(\widetilde{z}_{n+1},\mathcal{Z}^{tr})$. Without loss of generality, assume these scores are discrete-valued. For each $1\le i\le n_2^\prime$,
	\begin{equation*}
		\mathbb{P}\{\widetilde{S}_{n+1}=s_i|E_{\widetilde{z}}\}=\mathbb{P}\{\widetilde{Z}_{n+1}=z_i|E_{\widetilde{z}}\}=\frac{\sum_{\sigma:\sigma(n+1)=i}f(z_{\sigma(1)},\ldots,z_{\sigma(n+1)})}{\sum_{\sigma}f(z_{\sigma(1)},\ldots,z_{\sigma(n+1)})}
	\end{equation*}
	where $\sigma$ corresponds to the permutation of numbers $\{1,\ldots,n_2^\prime,n+1\}$ and $f$ is the joint density of $\{Z_1,\ldots,$ $Z_{n_2^\prime},\widetilde{Z}_{n+1}\}$. For instance, suppose $\sigma(i)=n+1$, then we have $z_{\sigma(i)}=\widetilde{z}_{n+1}$.  Notice that the condition distribution of $\widetilde{Y}_{n+1}$ given $\widetilde{X}_{n+1}$ is the same as those in the calibration data. 
	It follows from Lemma 1 that
	\begin{equation}\label{prob:v}
		\mathbb{P}\{\widetilde{S}_{n+1}=s_i|E_{\widetilde{z}}\}=\frac{\sum_{\sigma:\sigma(n+1)=i}\widehat{w}(x_{\sigma(n+1)})}{\sum_{\sigma}\widehat{w}(x_{\sigma(n+1)})}=\frac{\widehat{w}(x_{i})}{\sum_{i\in\mathcal{Z}^{cal,s}}\widehat{w}(x_{i})+\widehat{w}(\widetilde{x}_{n+1})}\equiv \widehat{p}_{i}(\widetilde{x}_{n+1}),
	\end{equation}
	Similarly, $$\mathbb{P}\{\widetilde{S}_{n+1}=\widetilde{s}_{n+1}|E_{\widetilde{z}}\}=\frac{\widehat{w}(x_{i})}{\sum_{i\in\mathcal{Z}^{cal,s}}\widehat{w}(x_{i})+\widehat{w}(\widetilde{x}_{n+1})}\equiv \widehat{p}_{n+1}(\widetilde{x}_{n+1}).$$ This yields that 
	\begin{equation*}
		\widetilde{S}_{n+1}|E_{\widetilde{z}}\sim \sum_{i=1}^{n_2^\prime}\widehat{p}_i(\widetilde{x}_{n+1})\delta_{s_{i}}+\widehat{p}_{n+1}(\widetilde{x}_{n+1})\delta_{\widetilde{S}_{n+1}}.
	\end{equation*}
	By Lemma 1 in \cite{tibshirani2019conformal}, it is equivalent to 
	\begin{equation}\label{wcp2}
		\widetilde{S}_{n+1}|E_{\widetilde{z}}\sim		
		\sum_{i=1}^{n_2^\prime}\widehat{p}_i(\widetilde{x}_{n+1})\delta_{v_{i}}+\widehat{p}_{n+1}(\widetilde{x}_{n+1})\delta_{\infty}.
	\end{equation}
	As a consequence, 
	\begin{eqnarray}
		&&\mathbb{P}(\widetilde{Y}_{n+1}\in\widehat{C}(\widetilde{X}_{n+1})|\mathcal{Z}^{tr})=\mathbb{P}(\widetilde{S}_{n+1}\le\eta(\widetilde{X}_{n+1})|\mathcal{Z}^{tr})\notag\\
		&&=\mathbb{P}(\widetilde{S}_{n+1}\le \text{Quantile}(1-\alpha;\sum_{i=1}^{n_2^\prime}\widehat{p}_i(\widetilde{x}_{n+1})\delta_{v_{i}}+\widehat{p}_{n+1}(\widetilde{x}_{n+1})\delta_{\infty}|\mathcal{Z}^{tr}))\ge 1-\alpha,\notag
	\end{eqnarray}
	where the last inequality follows from \eqref{wcp2}. In addition, we have that
	\begin{eqnarray}
		&&\big|\mathbb{P}(Y_{n+1}\in\widehat{C}(X_{n+1})|\mathcal{Z}^{tr},\mathcal{Z}^{cal})-\mathbb{P}(\widetilde{Y}_{n+1}\in\widehat{C}(\widetilde{X}_{n+1})|\mathcal{Z}^{tr},\mathcal{Z}^{cal})\big|\notag\\
		&&\le d_{\textrm{TV}}(\widetilde{P}_{X}\times{P}_{Y|\widetilde{A}=T,X}, P_{X}\times P_{Y^{\pi^e}|X}).\notag
	\end{eqnarray}
	Taking expectation with respect to $\mathcal{Z}^{cal}$ on both sides, we obtain that
	\begin{equation*}
		\big|\mathbb{P}(Y_{n+1}\in\widehat{C}(X_{n+1})|\mathcal{Z}^{tr})-\mathbb{P}(\widetilde{Y}_{n+1}\in\widehat{C}(\widetilde{X}_{n+1})|\mathcal{Z}^{tr})\big|\le d_{\textrm{TV}}(\widetilde{P}_{X}\times{P}_{Y|\widetilde{A}=T,X}, P_{X}\times P_{Y^{\pi^e}|X}).
	\end{equation*}
	Recall that $\widetilde{P}_{X}$ is defined as the distribution function such that $d\widetilde{P}_{X}(x)$ is proportional to $\widehat{w}(x)d{P}_{X|\widetilde{A}=T}(x)$. It follows that
	\begin{eqnarray}
		&&d_{\textrm{TV}}(\widetilde{P}_{X}\times{P}_{Y|\widetilde{A}=T,X}, P_{X}\times P_{Y^{\pi^e}|X})\notag\\
		&\le&\frac{1}{2}\sum_{t=0}^{m-1}\int\left|\widehat{w}(x)\frac{\pi_e(t|x)\pi_b(t|x)/\widehat{\pi}_b(t|x)}{\sum_{t^\prime}\pi_e(t^\prime|x)\pi_b(t^\prime|x)/\widehat{\pi}_b(t^\prime|x)}d{P}_{X|\widetilde{A}=T}(x)-w(x)\pi_e(t|x)dP_{X|A=T}(x)\right|\notag\\
		&+&\frac{1}{2}\sum_{t=0}^{m-1}\int\left|\widehat{w}(x)\frac{\pi_e(t|x)\pi_b(t|x)/\widehat{\pi}_b(t|x)}{\sum_{t^\prime}\pi_e(t^\prime|x)\pi_b(t^\prime|x)/\widehat{\pi}_b(t^\prime|x)}d{P}_{X|\widetilde{A}=T}(x)\left(1-\frac{1}{\int \widehat{w}(x)dP_{X|A=T}(x) }\right)\right| \notag \\
		&\overset{(1)}{=}&\frac{1}{2}\sum_{t=0}^{m-1}\int\left|{\pi_e(t|x)\pi_b(t|x)/\widehat{\pi}_b(t|x)}d{P}_{X}(x)-\pi_e(t|x)dP_{X}(x)\right|\notag\\
		&+& \frac{1}{2}\sum_{t=0}^{m-1}\int\left|{\pi_e(t|x)\pi_b(t|x)/\widehat{\pi}_b(t|x)}d{P}_{X}(x) \left(1-\frac{1}{\int \sum_{t=0}^{m-1} \pi_e(t|x)\pi_b(t|x)/\widehat{\pi}_b(t|x)dP_X(x) }\right) \right|  \notag\\
		&= &\frac{1}{2}\sum_{t=0}^{m-1}\int\pi_e(t|x)\pi_b(t|x)\left|{1/\widehat{\pi}_b(t|x)}-1/\pi_b(t|x)\right|dP_{X}(x)+\frac{1}{2}\sum_{t=0}^{m-1}\int{\pi_e(t|x)\pi_b(t|x)/\widehat{\pi}_b(t|x)}d{P}_{X}(x)-\frac{1}{2}\notag\\
		&{\le}&\sum_{t=0}^{m-1}\int\left|{1/\widehat{\pi}_b(t|x)}-1/\pi_b(t|x)\right|dP_{X}(x)\overset{(2)}{\rightarrow} 0,\notag
	\end{eqnarray}
	where (1) follows from \eqref{eqn:condx} and \eqref{eqn:weight} and (2) follows from Assumption \ref{ass:ss1}.

\end{proof}

\begin{proof}[\textbf{Proof of Theorem 2 (Exact coverage)}]
	The 
	proof is very similar to 
	that of Theorem 1 in \cite{tibshirani2019conformal} and is hence omitted.
\end{proof}

\begin{assumption}\label{ass:3}{(consistency of quantile regression estimates).} For $n_1$ large enough,
	\begin{eqnarray}
		&&\mathbb{P}\left[\mathbb{E}\left[(\widehat{q}_{\alpha_L}(X_{n+1})-{q}_{\alpha_L}(X_{n+1}))^{2}|\widehat{q}_{\alpha_L},\widehat{q}_{\alpha_U}\right]\le \eta_{n_1}\right]\ge 1-\rho_{n_1},\notag\\
		&&\mathbb{P}\left[\mathbb{E}\left[(\widehat{q}_{\alpha_U}(X_{n+1})-{q}_{\alpha_U}(X_{n+1}))^{2}|\widehat{q}_{\alpha_L},\widehat{q}_{\alpha_U}\right]\le \eta_{n_1}\right]\ge 1-\rho_{n_1},\notag
	\end{eqnarray}
	for some sequences $\eta_{n_1}=o(1)$ and $\rho_{n_1}=o(1)$ as $n_1\rightarrow\infty$.
\end{assumption}

\begin{proof}[\textbf{Proof of Theorem 3 (Asymptotic efficiency)}] The proof is similar to that of Theorem 1 in \cite{sesia2020comparison}. 
	First, we rewrite Assumption \ref{ass:3} as
	\begin{eqnarray}
		&&\mathbb{P}\left[\mathbb{E}\left[(\widehat{q}_{\alpha_L}(X_{n+1})-{q}_{\alpha_L}(X_{n+1}))^{2}|\widehat{q}_{\alpha_L},\widehat{q}_{\alpha_U}\right]\le {\eta_{n_1}}/{2}\right]\ge 1-{\rho_{n_1}}/{2},\notag\\
		&&\mathbb{P}\left[\mathbb{E}\left[(\widehat{q}_{\alpha_U}(X_{n+1})-{q}_{\alpha_U}(X_{n+1}))^{2}|\widehat{q}_{\alpha_L},\widehat{q}_{\alpha_U}\right]\le {\eta_{n_1}}/{2}\right]\ge 1-{\rho_{n_1}}/{2},\notag
	\end{eqnarray}
	for some sequences $\eta_{n_1}=o(1)$ and $\rho_{n_1}=o(1)$ as $n\rightarrow\infty$. Recall that 
	our prediction band is defined as
	\begin{equation*}
		\widehat{C}(X_{n+1})=[\widehat{q}_{\alpha_L}(X_{n+1})-Q_{1-\alpha}(X_{n+1}),\widehat{q}_{\alpha_U}(X_{n+1})+Q_{1-\alpha}(X_{n+1})]
	\end{equation*}
	where $Q_{1-\alpha}(X_{n+1})=(1-\alpha)\text{th quantile of}\textstyle\sum_{i\in\mathcal{Z}^{cal,s}}p_{i}(X_{n+1})\delta_{V_{i}}+p_{\infty}(X_{n+1})\delta_{\infty}$. The oracle band is given by
	\begin{equation*}
		C_{\alpha}^{oracle}(X_{n+1})=[{q}_{\alpha_L}(X_{n+1}),{q}_{\alpha_U}(X_{n+1})].
	\end{equation*}
	It suffices to show 
	\begin{eqnarray}
		(i)&& |\widehat{q}_{\beta}(X_{n+1})-{q}_{\beta}(X_{n+1})|=o_{P}(1) \text{~for~}\beta= \alpha_L,\alpha_U.\label{eqn:cond1}\\
		(ii)&&|Q_{1-\alpha}(X_{n+1})|=o_{P}(1)\label{eqn:cond2}
	\end{eqnarray}
	(i) Define random sets
	\begin{equation*}
		B_{n,U}=\{x:|\widehat{q}_{\alpha_U}(x)-q_{\alpha_U}(x)|\ge \eta_{n_1}^{1/3}\}, B_{n,L}=\{x:|\widehat{q}_{\alpha_L}(x)-q_{\alpha_L}(x)|\ge \eta_{n_1}^{1/3}\}
	\end{equation*}
	and $B_{n}=B_{n,U}\cup B_{n,L}$. we have
	\begin{eqnarray}
		&&\mathbb{P}[X_{n+1}\in B_{n}|\widehat{q}_{\alpha_L},\widehat{q}_{\alpha_U}]\notag\\
		&\le& \mathbb{P}[|\widehat{q}_{\alpha_U}(x)-q_{\alpha_U}(x)|^{2}\ge \eta_{n_1}^{2/3}|\widehat{q}_{\alpha_U}]+\mathbb{P}[|\widehat{q}_{\alpha_L}(x)-q_{\alpha_L}(x)|^{2}\ge \eta_{n_1}^{2/3}|\widehat{q}_{\alpha_L}]\notag\\
		&\le& \eta_{n_1}^{-2/3}\mathbb{E}[|\widehat{q}_{\alpha_U}(x)-q_{\alpha_U}(x)|^{2}]+\eta_{n_1}^{-2/3}\mathbb{E}[|\widehat{q}_{\alpha_L}(x)-q_{\alpha_L}(x)|^{2}]\le\eta_{n_1}^{1/3}\notag
	\end{eqnarray}
	with probability at least $1-\rho_{n_1}$ by Assumption \ref{ass:3}. This implies \eqref{eqn:cond1}.
	
	(ii) Consider the following partition of the data in $\mathcal{Z}^{cal,s}$, where $n_2^\prime=|\mathcal{Z}^{cal,s}|$:
	\begin{equation*}
		\mathcal{Z}^{cal,s}_{a}=\{i\in\mathcal{Z}^{cal,s}: Z_{i}\in B_{n}^{c}\},~\mathcal{Z}^{cal,s}_{b}=\{i\in\mathcal{Z}^{cal,s}: Z_{i}\in B_{n}\}.
	\end{equation*}
	First, by Hoeffding's inequality, 
	\begin{equation*}
		\mathbb{P}\left[|\mathcal{Z}^{cal,s}_{b}|\ge n_{2}^\prime\eta_{n_1}^{1/3}+t\right]\le \mathbb{P}\left[\frac{1}{n_{2}^\prime}\textstyle\sum_{i\in\mathcal{Z}^{cal,s}}\mathbb{I}[Z_{i}\in B_{n}]\ge \mathbb{P}[Z_{i}\in B_{n}]+\frac{t}{n_{2}^\prime}\right]\le\exp\left(-\frac{2t^{2}}{n_{2}^\prime}\right).
	\end{equation*}
	Set $t=c\sqrt{n_2^\prime\log n_2^\prime}$, 
	we obtain that $|\mathcal{Z}^{cal,s}_{b}|=o_{P}(n_2^\prime)=o_{P}(n)$.
	
	Next, define $\widetilde{S}_{i}=\max\{q_{\alpha_L}(X_i)-Y_i,Y_i-q_{\alpha_U}(X_i)\}$ for any $i\in\mathcal{Z}^{cal,s}$. By definition, for all $i\in\mathcal{Z}^{cal,s}_{a}$,
	\begin{equation}\label{eqn:Vi_Vi}
		\widetilde{S}_{i}-\eta_{n_1}^{1/3}\le S_{i}\le \widetilde{S}_{i}+\eta_{n_1}^{1/3}.
	\end{equation}
	Let $F_{n}$ and $\widetilde{F}_{n}$ denote the empirical distribution  $\textstyle\sum_{i\in\mathcal{Z}^{cal,s}}p_{i}(X_{n+1})\delta_{S_{i}}+p_{\infty}(X_{n+1})\delta_{\infty}$ and $\textstyle\sum_{i\in\mathcal{Z}^{cal,s}}p_{i}(X_{n+1})\delta_{\widetilde{S}_{i}}+p_{\infty}(X_{n+1})\delta_{\infty}$ respectively. Define $F_{n,a}$ and $\widetilde{F}_{n,a}$ as versions of $F_{n}$ and $\widetilde{F}_{n}$ when restricting attentions to observations that belong to $\mathcal{Z}^{cal,s}_{a}$ only. For sufficiently large $n$, we can show  $|\mathcal{Z}^{cal,s}_{b}|/|\mathcal{Z}^{cal,s}_{a}|\le\alpha$ by noting that $|\mathcal{Z}^{cal,s}_{b}|=o_{P}(n)$. 
	We next show that
	\begin{equation*}
		F_{n,a}^{-1}\left(1-\frac{n_{2}^\prime\alpha}{|\mathcal{Z}^{cal,s}_{a}|}\right)\le F_{n}^{-1}(1-\alpha)\le F_{n,a}^{-1}\left(1-\frac{n_{2}^\prime\alpha-|\mathcal{Z}^{cal,s}_{b}|}{n_2^\prime|\mathcal{Z}^{cal,s}_{a}|}\right).
	\end{equation*}
	To prove the first inequality, notice that for those observations that belong to $\mathcal{Z}^{cal,s}_{b}$, if their scores are in the lower $1-\alpha$ quantile of $F_{n}$, $F_{n,a}^{-1}(1-n_{2}^\prime\alpha/|\mathcal{Z}^{cal,s}_{a}|)=F_{n}^{-1}(1-\alpha)$. However, in general, the quantiles of $F_{n,a}$ will be smaller. The second inequality can be proven in a similar manner.  
	Combining this together with \eqref{eqn:Vi_Vi} yields that
	\begin{equation*}
		\widetilde{F}_{n,a}^{-1}\left(1-\frac{n_{2}^\prime\alpha}{|\mathcal{Z}^{cal,s}_{a}|}\right)-\eta_{n_1}^{1/3}\le F_{n}^{-1}(1-\alpha)\le \widetilde{F}_{n,a}^{-1}\left(1-\frac{n_{2}^\prime\alpha-|\mathcal{Z}^{cal,s}_{b}|}{n_2^\prime|\mathcal{Z}^{cal,s}_{a}|}\right)+\eta_{n_1}^{1/3}.
	\end{equation*}
	It in turn yields that $|\widetilde{F}_{n,a}^{-1}(1-\alpha)-\widetilde{F}_{n}^{-1}(1-\alpha)|=o_{P}(1)$ and hence,  $|\widetilde{F}_{n}^{-1}(1-\alpha)-F_{n}^{-1}(1-\alpha)|=o_{P}(1)$. By definition, $\widetilde{F}_{n}^{-1}(1-\alpha)=o_{P}(1)$. It follows that $Q_{1-\alpha}(X_{n+1}):={F}_{n}^{-1}(1-\alpha)=o_{P}(1)$. This yields \eqref{eqn:cond2}.
\end{proof}

\begin{proof}[\textbf{Proof of Theorem 4 (COPP-IS)}]
	%
	
	We index the data points in the calibration dataset $\mathcal{Z}^{cal}$ by $\{1,2,\ldots,n_2\}$. 
	Recall that $Z_i=(X_i,T_i,Y_i)$ for any $i\in\mathcal{Z}^{cal}$. Let $\widetilde{Z}_{n+1}=(\widetilde{X}_{n+1},\widetilde{E}_{n+1},\widetilde{Y}_{n+1})$ where $(\widetilde{X}_{n+1},\widetilde{Y}_{n+1})\sim \widetilde{P}_{X}\times{P}_{Y|\widetilde{A}=T,X}$ and $\widetilde{E}_{n+1}$ is the latent treatment variable. Let $E_{\widetilde{z}}$ denote the event that $\{Z_1,\ldots,Z_{n},\widetilde{Z}_{n+1}\}=\{z_1,\ldots,z_{n},\widetilde{z}_{n+1}\}$. Notice that each $Z_i$ involves the binary treatment variable. The corresponding nonconformity scores are denoted by $\widetilde{S}=(S_1,\ldots,S_{n},\widetilde{S}_{n+1})$ and $s_{i}=\mathcal{S}((x_i,y_i),\mathcal{Z}^{tr})$ for $1\le i\le n_2$, $\widetilde{s}_{n+1}=\mathcal{S}((\widetilde{x}_{n+1},\widetilde{y}_{n+1}),\mathcal{Z}^{tr})$. Notice that these scores are independent of the binary variable.

	Similar to the proof of Theorem 1, for each $1\le i\le n_2$,	
	%
	\begin{eqnarray}
		&&\mathbb{P}\{\widetilde{S}_{n+1}=v_i|E_{\widetilde{z}},A_1,\ldots,A_{n_2}\}\notag\\
		&=&\mathbb{P}\{(\widetilde{X}_{n+1},\widetilde{Y}_{n+1})=(x_i,y_i)|E_{\widetilde{z}},A_1,\ldots,A_{n_2}\}\notag\\
		&=&\frac{\widehat{w}(z_{i})I(A_i=t_i)}{\sum_{i\in\mathcal{Z}^{cal}}\widehat{w}(z_{i})I(A_i=t_i)+\widehat{w}(\widetilde{z}_{n+1})}=\widehat{p}_{i}(\widetilde{x}_{n+1}|A_1,\ldots,A_{n_2}).\notag
	\end{eqnarray}
	The quantile $Q_{1-\alpha}(\widetilde{x}_{n+1}|A_1,\ldots,A_{n_2})$ used to construct the interval is defined as $$(1-\alpha)\mbox{th quantile of}\sum_{i\in\mathcal{Z}^{cal}}\widehat{p}(\widetilde{x}_{n+1}|A_1,\ldots,A_{n_2})\delta_{V_i}+\widehat{p}_{n+1}(\widetilde{x}_{n+1}|A_1,\ldots,A_{n_2})\delta_{\infty}.$$
	In COPP-IS, 
	\begin{eqnarray}
		\mathbb{P}\{\widetilde{S}_{n+1}=v_i|E_{\widetilde{z}}\}
		=\frac{\widehat{w}(z_{i})\widehat{\pi}_{A}(t_i|x_i)}{\sum_{i\in\mathcal{Z}^{cal}}\widehat{w}(z_{i})\widehat{\pi}_{A}(t_i|x_i)+\widehat{w}(\widetilde{z}_{n+1})}=\widehat{p}_i(\widetilde{x}_{n+1}).\notag
	\end{eqnarray}
	The quantile $Q_{1-\alpha}(\widetilde{x}_{n+1})$ used to construct the interval is defined as $$(1-\alpha)\mbox{th quantile of}\sum_{i\in\mathcal{Z}^{cal}}\widehat{p}(\widetilde{x}_{n+1})\delta_{V_i}+\widehat{p}_{n+1}(\widetilde{x}_{n+1})\delta_{\infty}.$$
	
	By law of large numbers, $\lim_{n_2\rightarrow\infty}\{Q_{1-\alpha}(\widetilde{x}_{n+1}|A_1,\ldots,A_{n_2})-Q_{1-\alpha}(\widetilde{x}_{n+1})\}=0$. Other results can be similarly proven. 
	
\end{proof}

\begin{proof}[\textbf{Proof of Theorem 5 (COPP-MS)}] By Markov inequality, we have
	\begin{eqnarray}
		&&\mathbb{P}_{(X,Y^{\pi^e})\sim P_{X}\times P_{Y^{\pi^e}|X}}(Y^{\pi^e}\notin \widehat{C}_{B,\gamma}(X))\notag\\
		&=&\mathbb{P}_{(X,Y^{\pi^e})\sim P_{X}\times P_{Y^{\pi^e}|X}}(
		\sum_{b=1}^{B} \mathbb{I}(Y^{\pi^e}\notin \widehat{C}^{b}_{n_1,n_2}(X))\ge\gamma B)\notag\\
		&=&\mathbb{E}[\mathbb{I}(
		\sum_{b=1}^{B}\mathbb{I}(Y^{\pi^e}\notin \widehat{C}^{b}_{n_1,n_2}(X))\ge\gamma B)]\notag\\
		&\le& \frac{1}{\gamma B}\mathbb{E}[\sum_{b=1}^{B}\mathbb{I}(Y^{\pi^e}\notin \widehat{C}^{b}_{n_1,n_2}(X))]\notag\\
		&=&\frac{1}{\gamma}\mathbb{P}_{(X,Y^{\pi^e})\sim P_{X}\times P_{Y^{\pi^e}|X}}(Y^{\pi^e}\notin \widehat{C}^{b}_{n_1,n_2}(X)).\notag
	\end{eqnarray}
	According to Theorem 1, we have
	
	\begin{equation*}
		\lim_{n_1,n_1^\prime\rightarrow\infty}\mathbb{P}_{(X,Y^{\pi^e})\sim P_{X}\times P_{Y^{\pi^e}|X}}(Y^{\pi^e}\notin\widehat{C}_{n_1,n_2}^{b}(X))\le \alpha\gamma.
	\end{equation*}
	Combining the two results, we obtain that
	\begin{eqnarray}
		&&\lim_{n_1,n_1^\prime\rightarrow\infty}\mathbb{P}_{(X,Y^{\pi^e})\sim P_{X}\times P_{Y^{\pi^e}|X}}(Y^{\pi^e}\notin\widehat{C}_{B,\gamma}(X))\notag\\
		&\le& \lim_{n_1,n_1^\prime\rightarrow\infty}\frac{1}{\gamma}\mathbb{P}_{(X,Y^{\pi^e})\sim P_{X}\times P_{Y^{\pi^e}|X}}(Y^{\pi^e}\notin \widehat{C}^{b}_{n_1,n_2}(X))\le\alpha.\notag
	\end{eqnarray}
	This is equivalent to 
	\begin{equation*}
		\lim_{n_1,n_1^\prime\rightarrow\infty}\mathbb{P}_{(X,Y^{\pi^e})\sim P_{X}\times P_{Y^{\pi^e}|X}}(Y^{\pi^e}\in\widehat{C}_{n_1,n_2}^{b}(X))\ge 1-\alpha.
	\end{equation*}
	The proof is hence completed. 
\end{proof}

\section{SEQUENTIAL DECISION MAKING}
This section provides assumptions, pseudo codes and theories 
in sequential decision making. Finally, we conclude this section with a discussion to extend our proposal to settings with 
immediate rewards at each decision point.
We begin with the consistency, sequential ignorability and positive assumption in sequential decision making. Let $Y_{i}{(t_1,t_2,\ldots,t_{K})}$ denote the reward that the $i$th instance would be observed were they to receive action $t_1,t_2,\ldots,t_K$ sequentially. The standard assumptions are (1) $Y_{i}{(T_{1i},T_{2i},\ldots,T_{Ki})}=Y_{i}$ almost surely for any $i$ (i.e., consistenct); (2) A policy $\pi_b$ satisfies sequential ignorability, that is at any stage $k$, conditional on the history $H_{k}$ generated by the policy, the action $T_{k}$ generated by the policy is independent of the potential outcomes $\{X_{k+1}(t_1,\ldots,t_{k}), X_{k+2}(t_1,\ldots,t_{k+1}),\ldots X_{K}(t_1\ldots,t_{K-1}),Y(t_{1},t_{2},\ldots,t_{K})\}$ for all $t_{k}\in\{0,1,\ldots,m-1\}$. (3) $\pi_{b_{k}}(t_{k}|h_{k})$ is uniformly bounded away from zero for any $t_{k},h_{k}$ (i.e., positivity).

Denote $\mathbf{A}=(A_1,\ldots,A_K)^\top$ as the actions generated by the pseudo policy with $\pi_{a_{k}}(t_k|h_k)\propto\pi_{e_k}(t_k|h_k)/\pi_{b_{k}}(t_k|h_k)$ and $\mathbf{T}=(T_1,\ldots,T_K)^\top$ as the treatment generated by the behavior policy, and 
$\widetilde{\textbf{A}}$ as the one generated by the estimated pseudo policy 
$\pi_{\widetilde{a}_k}(t_k|h_k)\propto\pi_{e_k}(t_k|h_k)/\widehat{\pi}_{b_k}(t_k|h_k)$.
We summarize the pseudocode of our proposal COPP in Algorithm \ref{alg2}.

\begin{algorithm}[ht]
	\caption{COPP: Conformal off-policy prediction in multi-stage decision making}\label{alg2}
	\begin{algorithmic}[t]
		\State \textbf{Input:}
		Data $\{(X_{1i},T_{1i},\ldots,X_{Ki},T_{Ki},Y_{i})\}_{i=1}^{n}$; a test point with initial state $X_{1,n+1}$; a 
		sequence of target policies $\pi_e=\{\pi_{e_k}(t_{k}|h_k)\}_{k=1}^{K}$; propensity score training algorithm $\mathcal{P}$; quantile prediction algorithm $\mathcal{Q}$; conformity score $\mathcal{S}$; and coverage level $1-\alpha$ with $\alpha_U-\alpha_L=1-\alpha$.
		\begin{itemize}
			\item[1:] Split the data into two disjoint subsets $\mathcal{Z}^{tr}$ and $\mathcal{Z}^{cal}$.
			\item[2:] Train $\{\widehat{\pi}_{b_k}(t_k|h_{k})\}_{k=1}^{K}$ using $\mathcal{P}$ on all samples from $\mathcal{Z}^{tr}$, i.e., $$\widehat{\pi}_{b_k}(t_k|h_k)\leftarrow \mathcal{P}(\{(H_{ki},T_{ki})\}_{i\in\mathcal{Z}^{tr}}), H_{ki}=\{(X_{1i},T_{1i},\ldots,X_{ki})\},\mbox{~for~}1\le k\le K.$$
			\item[3:] Draw $\widetilde{\mathbf{A}}_i=(A_{1i},\ldots,A_{Ki})$ for $i=1,\ldots,n$ with plugging $\widehat{\pi}_{b_k}(t_k|h_k)$.
			\item[\textbf{4:}] Train $\widehat{w}(X_1)$ using $\mathcal{P}$ on all samples from the  $\mathcal{Z}^{tr}$ augmented by $\{\widetilde{\mathbf{A}}_{i}\}_{i\in\mathcal{Z}^{tr}}$, i.e.,
			$$\widehat{e}(X_1)=\widehat{\mathbb{P}}(\widetilde{\mathbf{A}}=\mathbf{T}|X_1,\mathcal{Z}^{tr})\leftarrow \mathcal{P}(\{(\widetilde{\mathbf{A}}_i,\mathbf{T}_i,X_{1i})\}_{i\in\mathcal{Z}^{tr}}), \widehat{w}(X_1)=1/\widehat{e}(X_1).$$
			\item[5:] Select subsamples satisfying $\widetilde{\mathbf{A}}_i=\mathbf{T}_i$ in both subsets denoted as $\mathcal{Z}^{tr,s}$ and $\mathcal{Z}^{cal,s}$. 
			\item[6:] Train quantile regressions using $\mathcal{Q}$ on selected subsamples from $\mathcal{Z}^{tr,s}$, i.e., $$\widehat{q}_{\alpha_L}(x;\mathcal{Z}^{tr,s})\leftarrow \mathcal{Q}(\alpha_L,\{(X_{1i},Y_{i})\}_{i\in\mathcal{Z}^{tr,s}}), \widehat{q}_{\alpha_U}(x;\mathcal{Z}^{tr,s})\leftarrow \mathcal{Q}(\alpha_U,\{(X_{1i},Y_{i})\}_{i\in\mathcal{Z}^{tr,s}}).$$
			\item[7:] Compute the nonconformity scores for all selected subsamples $i\in \mathcal{Z}^{cal,s}$: $$S_i=\max\{\widehat{q}_{\alpha_L}(X_{1i};\mathcal{Z}^{tr,s})-Y_i, Y_i-\widehat{q}_{\alpha_U}(X_{1i};\mathcal{Z}^{tr,s})\}.$$
			\item[8:] Compute the normalized weights for $i\in \mathcal{Z}^{cal,s}$ and the test point $X_{n+1}$
			{\small 
				$$\widehat{p}_{i}(X_{1,n+1})=\frac{\widehat{w}(X_{1i})}{\textstyle\sum_{i\in\mathcal{Z}^{cal,s}}\widehat{w}(X_{1i})+\widehat{w}(X_{1,n+1})},~ \widehat{p}_{\infty}(X_{1,n+1})=\frac{\widehat{w}(X_{1,n+1})}{\textstyle\sum_{i\in\mathcal{Z}^{cal,s}}\widehat{w}(X_{1i})+\widehat{w}(X_{1,n+1})}.$$}
			\item[9:] Compute the $(1-\alpha)$th quantile of $\sum_{i\in\mathcal{Z}^{cal, s}}\widehat{p}_i(X_{1,n+1})\delta_{S_{i}}+\widehat{p}_{\infty}(X_{1,n+1})\delta_{\infty}$ as $Q_{1-\alpha}(X_{1,n+1})$.
			\item[10:] Construct a prediction set for $X_{1,n+1}$:
			$$\widehat{C}(X_{1,n+1})=[\widehat{q}_{\alpha_L}(X_{1,n+1};\mathcal{Z}^{tr,s})-Q_{1-\alpha}(X_{1,n+1}),\widehat{q}_{\alpha_U}(X_{1,n+1};\mathcal{Z}^{tr,s})+Q_{1-\alpha}(X_{1,n+1})].$$
		\end{itemize}
		\State \textbf{Output:} A prediction set $\widehat{C}(X_{1,n+1})$ for the outcome $Y^{\pi^e}_{n+1}$. 
	\end{algorithmic}
\end{algorithm}

Let $n_1=|\mathcal{Z}^{tr}|$.
The following assumption provides the consistency of behavior policy estimates. 

\begin{assumption}\label{ass:SMD}
	For the output in Step 2 of Algorithm \ref{alg2} and any $1\le k\le K$, $\widehat{\pi}_{b_k}(t_k|h_k)$s are uniformly bounded away from zero and $$\lim_{n_1\rightarrow\infty}\mathbb{E}\big|{\widehat{\pi}_{b_k}(t_k|H_k)}-{{\pi}_{b_k}(t_k|H_k)}\big|=0.$$
\end{assumption}
\begin{assumption}\label{ass:SMD2}
	Suppose that the output $\widehat{\mathbb{P}}(\widetilde{\mathbf{A}}=\mathbf{T}|X_1,\mathcal{Z}^{tr})$ in Step 4 of Algorithm \ref{alg2} 
	is uniformly bounded away from zero and
	$$\lim_{n_1\rightarrow\infty}\mathbb{E}_{X_1\sim P_{X_1}}\big|\widehat{\mathbb{P}}(\widetilde{\mathbf{A}}=\mathbf{T}|X_1,\mathcal{Z}^{tr})-\mathbb{P}(\widetilde{\mathbf{A}}=\mathbf{T}|X_1,\mathcal{Z}^{tr})\big|=0.$$
\end{assumption}

Let ${P}_{X_1|\widetilde{\mathbf{A}}=\mathbf{T}}$ be the probability measure of the initial state for selected samples in $\mathcal{Z}^{cal, s}$ as an estimate of ${P}_{X_1|{\mathbf{A}}=\mathbf{T}}$. Denote $w(X_1)=1/\mathbb{P}(\mathbf{A}=\mathbf{T}|X_1)$ and $\widehat{w}(X_1)=1/\widehat{\mathbb{P}}(\widetilde{\mathbf{A}}=\mathbf{T}|X_1,\mathcal{Z}^{tr})$ as the output in Step 4 of Algorithm \ref{alg2}. We summarize the theoretical results below. 

\begin{theorem}[\textbf{Asymptotic coverage for SDM}]\label{thm:SDM}
	Let $n_1^\prime=|\mathcal{Z}^{tr,s}|$. Suppose that Assumptions \ref{ass:SMD}-\ref{ass:SMD2} hold and $\mathbb{E}_{X_1\sim P_{X_1}}[w(X_1)]<\infty$, $\mathbb{E}_{X_1\sim P_{X_1}}[\widehat{w}(X_1)|\mathcal{Z}^{tr}]<\infty$,  $\mathbb{E}_{X_1\sim P_{X_1|\widetilde{\mathbf{A}}=\mathbf{T}}}[\widehat{w}(X_1)|\mathcal{Z}^{tr}]<\infty$, then the output $\widehat{C}(x)$ from Algorithm \ref{alg2} satisfies
	\begin{equation}
		\lim_{n_1,n_1^\prime\rightarrow\infty}\mathbb{P}_{(X_1,Y^{\pi^e})\sim P_{X_1}\times P_{Y^{\pi^e}|X_1}}(Y^{\pi^e}\in\widehat{C}(X_1))\ge 1-\alpha,
	\end{equation}
\end{theorem}

\textbf{Discussion.} We conclude this section by extending our proposal to settings with immediate rewards at each decision point. Suppose the observed data can be summarized as $\{(X_{1i},T_{1i},Y_{1i},X_{2i},T_{2i},Y_{2i},\ldots,X_{Ki},T_{Ki},Y_{Ki})\}_{i=1}^{n}$ where $Y_{ki}$ is the immediate reward at the $k$th stage. Similarly, we can show that the conditional distribution $Y_{k}|A_1=T_1,\ldots,A_k=T_k,H_1$ is the same as that of $Y^{\pi^e}_{k}|H_1$ for each $1\le k\le K$. As such, for each $k$, we can apply our proposal to construct a PI for $Y^{\pi^e}_k$. These PIs can be potentially further aggregated to cover the sum $\sum_k Y^{\pi^e}_k$. We leave it for future research. 

\section{ADDITIONAL IMPLEMENTATION DETAILS}
This section provides additional implementation details for the competing methods in the simulation study. First, notice that the importance sampling (IS) method can be naturally coupled with the kernel method to evaluate the individual treatment effect (ITE). Specifically, consider the following estimator, 
\begin{equation*}
	\widehat{\mathbb{E}}[Y_{n+1}^{\pi_e}|X_{n+1}]=\sum_{i=1}^{n}\frac{\pi_e(T_i|X_i)}{\widehat{\pi}_b(T_i|X_i)}Y_i\frac{K((X_i-X_{n+1})/h)}{\sum_{i=1}^{n}K((X_i-X_{n+1})/h)},
\end{equation*}
for certain kernel function $K$ with bandwidth $h$, 
and 
the logistic regression estimator estimator $\widehat{\pi}_b$. Its standard deviation can be estimated based on the sampling variance estimator and the corresponding confidence interval (CI) can be derived. In our experiments, the bandwidth parameter $h$ is manually selected so that the resulting CI achieves the best empirical coverage rate. 

Second, the double robust (DR) estimator can be coupled with kernel method for ITE evaluation as well. Specifically, define
\begin{equation*}
	\widehat{\mathbb{E}}[Y_{n+1}^{\pi_e}|X_{n+1}]=\sum_{i=1}^{n}\left\{\frac{\pi_e(T_i|X_i)}{\widehat{\pi}_b(T_i|X_i)}[Y_i-\widehat{\mu}(X_{i})]+\widehat{\mu}(X_{i})\right\}\frac{K((X_i-X_{n+1})/h)}{\sum_{i=1}^{n}K((X_i-X_{n+1})/h)},
\end{equation*}
where $\widehat{\mu}(x)$ denotes the estimated regression function obtained via random forest. 
The corresponding confidence interval can be similarly constructed. 

\vfill

\end{document}